%% file: main.tex
\pdfoutput=1

\documentclass[11pt]{article}

\usepackage[final]{acl}

\usepackage{times}
\usepackage{latexsym}

\usepackage[T1]{fontenc}

\usepackage[utf8]{inputenc}

\usepackage{microtype}

\usepackage{inconsolata}

\usepackage{graphicx}
\usepackage{amsmath}
\usepackage{amssymb}
\usepackage{booktabs}
\usepackage{algorithm}
\usepackage{algpseudocode}
\usepackage[inkscapelatex=false]{svg}
\usepackage{multirow}
\usepackage{makecell}
\usepackage{pifont}
\usepackage{booktabs}
\usepackage{subcaption}
\usepackage{colortbl}
\usepackage{tabularx}  
\usepackage{array}  
\usepackage{caption}  
\usepackage{longtable}  
\usepackage{float}  

\colorlet{lviolet}{violet!15}
\colorlet{lblue}{blue!15}
\colorlet{lylw}{yellow!40}
\newcommand{\lvio}{\cellcolor{lviolet}}
\newcommand{\lblue}{\cellcolor{lblue}}
\newcommand{\lylw}{\cellcolor{lylw}}

\title{Mitigating Hallucinations in Vision-Language Models through Image-Guided Head Suppression}

\author{Sreetama Sarkar$^{1,}$\thanks{Equally contributing authors.} \ \ \ 
Yue Che$^{1, \ast}$ \ \ \ 
Alex Gavin$^{3}$ \ \ \
Peter A. Beerel$^{1}$\ \ \
Souvik Kundu$^{2}$ \\
$^{1}$University of Southern California, Los Angeles, USA \ \ \
$^{2}$Intel Labs, USA \\
$^{3}$Harvard-Westlake School, Los Angeles, USA \\
{\tt\small {\{sreetama,yueche,pabeerel\}@usc.edu} \ \  \tt\small {souvikk.kundu}@intel.com}}

\begin{document}
\maketitle
\begin{abstract}
Despite their remarkable progress in multimodal understanding tasks, large vision language models (LVLMs) often suffer from \textit{``hallucination''}, generating texts misaligned with the visual context. Existing methods aimed at reducing hallucinations through inference time intervention incur a significant increase in latency. To mitigate this, we present \textbf{SPIN}, a task-agnostic attention-guided head suppression strategy that can be seamlessly integrated during inference \textbf{\textit{without incurring any significant compute or latency overhead}}. We investigate whether hallucination in LVLMs can be linked to specific model components. Our analysis suggests that hallucinations can be attributed to a dynamic subset of attention heads in each layer. Leveraging this insight, for each text query token, we selectively suppress attention heads that exhibit low attention to image tokens, keeping the top-$k$ attention heads intact. Extensive evaluations on visual question answering and image description tasks demonstrate the efficacy of SPIN in reducing hallucination scores up to 
$\mathbf{2.7}\times$ while maintaining F1, and improving throughput by 
$\mathbf{1.8}\times$ compared to existing alternatives. Code is available \href{https://github.com/YUECHE77/SPIN}{here}.  
\end{abstract}

\section{Introduction}
 Large language models (LLMs) \cite{touvron2023llama} have revolutionized natural language understanding and generation, achieving state-of-the-art (SoTA) performance across numerous tasks. To extend these capabilities to vision-language tasks, large vision language models (LVLMs)~\cite{Liu_2024_llava1.5, zhu2024minigpt} integrate an LLM backbone with vision encoders, mapping image inputs into the text embedding space. While this approach has enabled remarkable progress in vision-language understanding, LVLMs often generate output text misaligned with the visual context, a phenomenon commonly referred to as  ``\textit{hallucinations}'', which undermines their reliability in critical domains such as healthcare, autonomous driving, and surveillance. 

\begin{figure}[t!]
    \centering
    \includegraphics[trim = 70 70 70 120, clip, width=\linewidth]{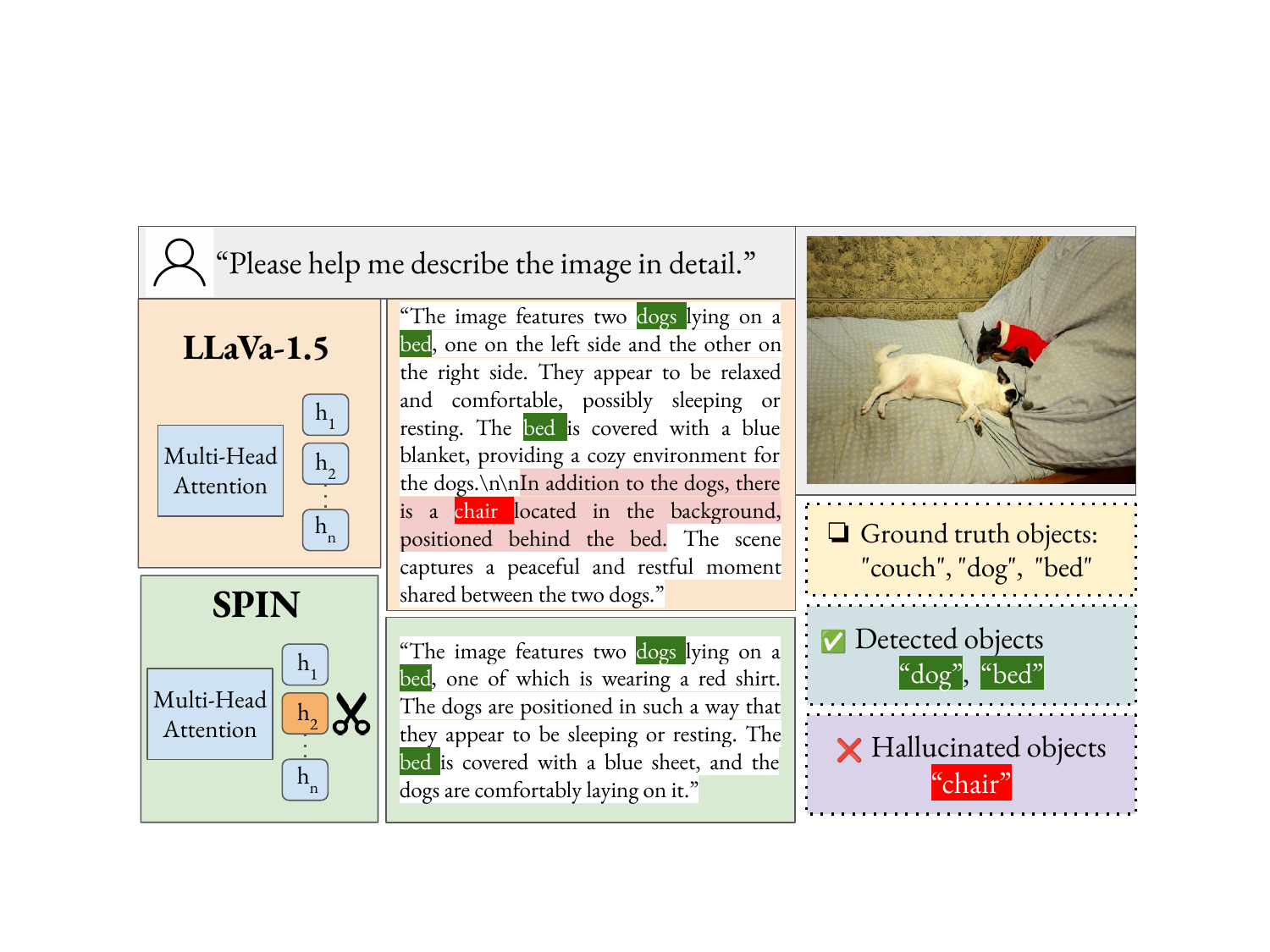}  
    \vspace{-6mm}
    \caption{Caption generation using LLaVA-1.5 and SPIN. LLaVA-1.5's generated text description mentions a ``chair'' in the background, which is clearly a hallucinated object. SPIN mitigates hallucination while successfully identifying the objects present in the image.}
    \vspace{-6mm}
    \label{fig:block_diagram}
\end{figure}
 
One key source of hallucination in LVLMs is textual bias, inherited from the pre-trained LLM backbone \cite{liu2024PAI, chen2024fastv}. To mitigate this, fine-tuning strategies based on Reinforcement Learning (RL) with human or AI-generated feedback have been proposed \cite{sun-etal-2024-mmhalbench, jing2025fgaif}. However, these methods are computationally intensive and often impractical for deployment in resource-constrained environments. Recent research has explored inference-only approaches, which modify the decoding pipeline \cite{huang2024opera} or apply contrastive decoding techniques \cite{leng2024vcd, liu2024PAI} to refine logit scores and reduce hallucinations. Although these methods are significantly more efficient compared to training-based approaches, they often struggle to sufficiently reduce hallucinations and suffer from significantly increased latency.

Besides, there is a lack of systematic understanding of how different model components contribute to hallucinations. In this paper, we investigate the role of attention heads in hallucinated text generation. We show that for each input token, a subset of attention heads in each layer disproportionately contributes to hallucinations. By identifying and suppressing these heads, we demonstrate that we can effectively reduce hallucinations in LVLMs while preserving SoTA model performance, as demonstrated in Figure \ref{fig:block_diagram}.


\vspace{1mm}
\noindent\textbf{Our Contributions:}
We perform a detailed analysis revealing that hallucination in LVLMs often stems from specific attention heads exhibiting insufficient attention to visual input. We characterize and quantify these "image-inattentive" heads across model layers. Based on this insight, we propose \textbf{SPIN}, \textbf{S}u\textbf{P}pressing image \textbf{IN}attentive heads, a novel, attention-guided head suppression strategy that can be seamlessly integrated during inference, irrespective of the decoding strategy or projection modules. SPIN offers a highly efficient solution since it achieves substantial hallucination reduction without requiring model retraining, and crucially, introduces \textbf{\textit{no additional computational overhead or latency during inference}}.
We evaluate our approach on visual question answering (VQA) and image caption generation tasks. SPIN reduces hallucination scores up to~$\mathbf{2.7}\times$ over existing methods, while improving throughput by up to $\mathbf{1.8}\times$.

\section{Background and Related Work}
\subsection{LVLM Preliminaries}
LVLMs typically comprise four key components: a text tokenizer, an image encoder, a projector, and a language decoder. The text tokenizer processes the language input by segmenting it into discrete tokens and converting them into text embeddings, henceforth referred to as \textit{text tokens}. Similarly, the image encoder partitions the input image into patches and transforms them into corresponding visual embeddings. These visual embeddings are then mapped into the text embedding space through the projector, producing \textit{vision tokens}. Finally, the vision and text tokens are concatenated and fed into the language decoder.

The language decoder consists of a series of transformer encoder blocks, each consisting of a multi-head attention (MHA) layer followed by a feed-forward network (FFN).  The MHA takes in $N$ input tokens, consisting of both text and vision tokens, each of embedding dimension $d$, \(X \in \mathbb{R}^{N \times d} \) and maps them into Query, Key and Value matrices \((Q, K, V) \in \mathbb{R}^{N \times d}\). MHA then computes low-dimensional projections of \((Q, K, V)\) given by \((Q^i, K^i, V^i) \in \mathbb{R}^{N \times d_k} \) for each head \(i\) where \(d_k = d/H\), $H$ denoting the number of attention heads. The scaled dot product attention for each head is then computed using 
\begin{equation}
    h_i = \text{Softmax}\left(\frac{Q^i K^{iT}}{\sqrt{d_k}}\right) V^i
\end{equation}
The results for each head are concatenated and a final projection matrix \(W_o \in \mathbb{R}^{d \times d}\) is applied to obtain the MHA output.
\begin{equation}
    \text{MHA}_{Q, K, V} = \left( \bigoplus_{i=1}^{H} h_i \right)  W_o
    \label{eqn:MHA}
\end{equation}

The output text sequence is generated based on a \emph{decoding strategy}, which determines how the next token is selected from the probability distribution obtained from the language decoder. Some of the common approaches include \emph{greedy search}, which deterministically selects the token with the highest probability at each step; \emph{beam-search} \cite{freitag-al-onaizan-2017-beam}, which explores multiple candidate token sequences ("beams") simultaneously to find a higher-scoring overall sequence; and \emph{nucleus sampling} \cite{nucleus_sampling}, which samples the next token randomly from the smallest set of tokens whose cumulative probability exceeds a given threshold. Additionally, a repetition penalty is often applied during decoding to discourage the model from generating repetitive text by reducing the likelihood of tokens that have recently appeared in the sequence.

\subsection{LVLM Hallucination and Mitigation}
Hallucinations can manifest as factual errors to a user's query or inaccurate image description resulting in non-existent objects, incorrect object attributes or relationships~\cite{liu2024survey}. 
Several causes of hallucinations have been identified including biased training data~\cite{liu2023mitigating}, the inability of vision encoders
to accurately ground images \cite{jain2024vcoder}, misalignment among different
modalities~\cite{Liu_2024_llava1.5}, and insufficient context attention in LLM decoders~\cite{huang2024opera, liu2024PAI}. 
Existing hallucination mitigation approaches can be broadly classified into \textit{training-based} and \textit{training-free} methods. 

\vspace{1mm}
\noindent\textbf{Training-based Mitigation:}
\cite{sun-etal-2024-mmhalbench} adapt the Reinforcement
Learning from Human Feedback (RLHF), originally developed for text-only models, to the vision-language setting, training VLMs to maximize simulated human rewards. They propose Factually-Augmented RLHF, where the reward model is given access to additional ground truth information, such as image captions, to improve its assessment of factual correctness.  In contrast, FGAIF \cite{jing2025fgaif} replaces human supervision with fine-grained AI-generated feedback. It segments model responses into sub-sentences and uses AI models to detect hallucinations relating to object existence, attributes, and relationships.
LACING~\cite{zhao2024looking} addresses language bias in LVLMs by introducing a multimodal dual-attention mechanism and soft-image guidance. It constructs separate attention streams for visual and textual inputs to improve grounding and alignment. Despite their effectiveness, training-based methods are computationally expensive, requiring large-scale resources (e.g., 8× A100 GPUs with 40GB memory each), making them impractical in resource-constrained settings.

\vspace{1mm}
\noindent\textbf{Training-free Mitigation:}
Recently, training-free approaches 
have been proposed to mitigate the issue of visual context neglect in LVLMs. OPERA\cite{huang2024opera} observes that hallucination arises from generating new tokens based on limited summary tokens in which attention aggregation occurs, leading the model to ignore the image context. To address this, OPERA introduces a beam-search variant with a weighted scoring mechanism that downranks candidate sequences exhibiting over-trust patterns. However, its applicability is restricted to beam-search, which is significantly slower than greedy decoding. Other techniques leverage contrastive decoding, which requires multiple forward passes, increasing latency.
VCD~\cite{leng2024vcd} builds on the observation that increased visual uncertainty drives models to rely on language priors, thereby amplifying hallucinations. It contrasts the output distributions conditioned on original and distorted visual inputs to identify and suppress hallucinated content.
PAI~\cite{liu2024PAI} mitigates the reduced attention to image tokens through a two-step approach: (1) amplifying attention to image tokens and (2) refining logit scores by subtracting logits computed without image prior to eliminate the text bias. 
DAMRO \cite{gong-etal-2024-damro} attributes high-attention outlier tokens scattered in the background of the image as the cause for hallucinations and proposes to mitigate the influence of outlier tokens using contrastive decoding.
HALC \cite{chen2024halc} is based on contrasting distributions with different visual contexts and using visual matching scores for candidate selection in beam-search. While all the above approaches introduce latency overhead, HALC is reported to cost around 2.4$\times$ of the normal greedy decoding time. Recently, \cite{liu2025vti} introduced a test-time latent space steering approach through visual and textual intervention (VTI) using pre-computed steering vectors, whereas ICT \cite{Chen_2025_ict} proposes image and object-level intervention to apply targeted activation shifts to selected attention heads identified through binary classifiers, requiring additional parameters.
In contrast, we propose a training-free mitigation strategy that introduces no additional overhead by identifying the role of attention heads in hallucination.

\subsection{Role of Attention Heads}
Earlier works
\cite{voita2019storyofheads} analyzed the role of individual attention heads in neural machine translation, showing that most heads can be pruned without significantly affecting model performance. More recently, routing to expert attention heads \cite{jin2024moh, zheng2025citer} was introduced for LLMs and vision transformers, allowing each input token to dynamically select appropriate heads, thereby improving performance. \cite{zhou2024ships} further highlighted the impact of specific attention heads on LLM safety, demonstrating that ablating a small subset of heads increases the attack success rate. Recently, \cite{kundu2025lvlm} explored the impact of LLM weight \cite{you2024shiftaddllm} and KV quantization \cite{kang2024gear} on accuracy as well as hallucination performance of LLMs.  However, the role of attention heads in the context of hallucination remains unexplored.  

\begin{figure}[t!]
    \centering
    \includegraphics[width=\linewidth]{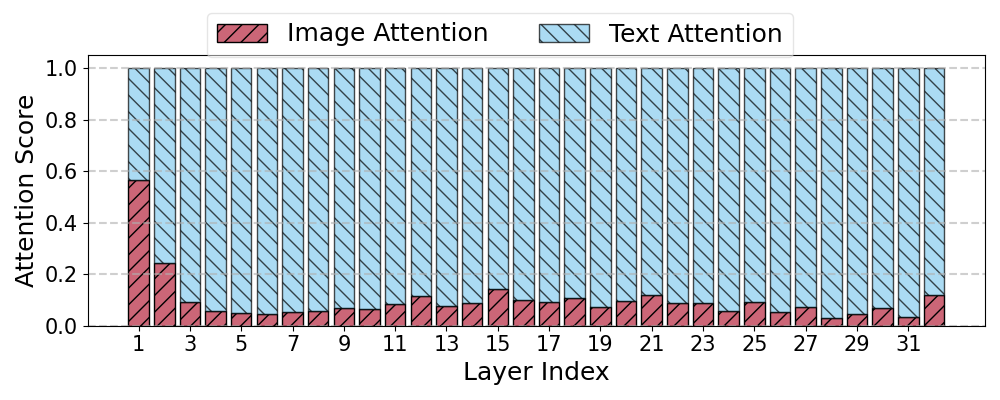}
    \vspace{-8mm}
    \caption{Average attention allocated by the current token to the preceding vision and text tokens in CHAIR. Image tokens receive~$<$10\% of total attention from layer 3, while constituting~$\sim$76-92\% of the input.}
    \vspace{-5mm}
    \label{fig:attention}
\end{figure}

\section{Methodology}
\noindent\textbf{Motivational Study:}
LVLMs struggle with inefficient image attention~\cite{chen2024fastv, liu2024PAI}. In Figure~\ref{fig:attention}, we illustrate how the current token distributes attention across preceding vision and text tokens (constituting system prompt, instruction prompt, and generated output) averaged over all output tokens in the image caption. Our analysis reveals that in deeper layers, generated text tokens allocate~$<$10\% of their attention to vision tokens, despite vision tokens comprising~$\sim$76-92\% of the input. This imbalance causes the model to ignore the context provided by the visual input (taken as the ``fact''), potentially leading to hallucinations. To mitigate this, for each query text token, we identify a fraction of heads that allocates the least cumulated attention to vision tokens. 
We then present a training-free strategy to dynamically suppress attention heads, reducing the imbalance in attention and thereby enhancing model performance.

\vspace{1mm}
\noindent\textbf{SPIN Multi-Head Attention:} We propose to mitigate problematic attention heads across layers using a dynamic mask \(m_i\) for each attention head \(i\). 
\( m_i \) is obtained based on the attention of the current text query token \( q^i \in \mathbb{R}^{1 \times d_k} \) to key vision tokens, having length \(N_v\) given by:
\begin{equation}
\text{A}_{v} = \text{A}_{tot}[I_{\text{start}}:I_{\text{end}}], \ \ \text{A}_{tot} = q^i K^{iT}
\end{equation}
Here, \(\text{A}_{tot} \in \mathbb{R}^{1 \times N}\) denotes the attention of \( q^i\) to \( N \) input tokens, \( K^i \in \mathbb{R}^{N \times d_k} \) represents the key matrix for \( N \) tokens, and \(\text{A}_{v} \in \mathbb{R}^{1 \times N_{v}}\) denotes the attention score for \( q^i\) to the vision tokens only. \(I_{\text{start}}\) and \(I_{\text{end}}\) denotes the start and end indices of vision tokens.
The mask \( m_i \) is defined as:
\[
m_i = 
\begin{cases}
1 & \text{if } i \in \text{top-$k$}(\sum_{j=1}^{N_v} A_v[j]) \\
\alpha & \text{otherwise}
\end{cases}
\]
Thus, \( m_i \) is set to 1, identifying that we should keep the head intact if the $i^{th}$ attention head belongs to the top $k$ highest values across $n$ heads. Otherwise, \( m_i \) is set to \( \alpha \), which is the suppression factor used to reduce the influence of the $i^{th}$ head. If \( \alpha = 0 \), the head is completely suppressed or effectively pruned. We denote the ratio of suppressed heads as $r$ where $r=(1-k/H)$. The choice of layers for suppressed heads, suppressed head ratio $r$ and suppression factor $\alpha$ are the key hyperparameters in our approach. The final multi-head attention in SPIN is computed as:
\begin{equation}
    \text{MHA}_{Q, K, V, m} = \left( \bigoplus_{i=1}^{H} (m_i \cdot h_i) \right) W_o.
\end{equation}

\noindent\textbf{Head and Suppression Factor Selection:} To identify the dynamic subset of problematic attention heads for a specific model and task, we adopt an efficient systematic three-stage approach. \emph{First,} we vary $r$ to find the value that achieves the best reduction in hallucinations without a significant drop ($\sim$3\%) in F1. Here, we prune attention heads across all layers equally setting $\alpha=0$. Adjusting these parameters is explored next.  
\emph{Second,} our analysis (Section~\ref{sec:ablation}) reveals that attention heads in earlier layers tend to contribute more to hallucinations than those in later layers. For specific models and tasks, we find the optimal number of early layers in which we should prune attention heads to minimize hallucination scores.
\emph{Finally,} we explore increasing $\alpha$ for the selected heads to mitigate the drop in F1 observed in step 1. In particular, a higher $\alpha$ results in improving F1 score at the cost of increasing hallucinations. We choose an $\alpha$ that provides an optimal trade-off between hallucinations and F1.

\section{Experimental Results}
\begin{figure*}[t!]
    \centering
      \includegraphics[width=0.90\linewidth]{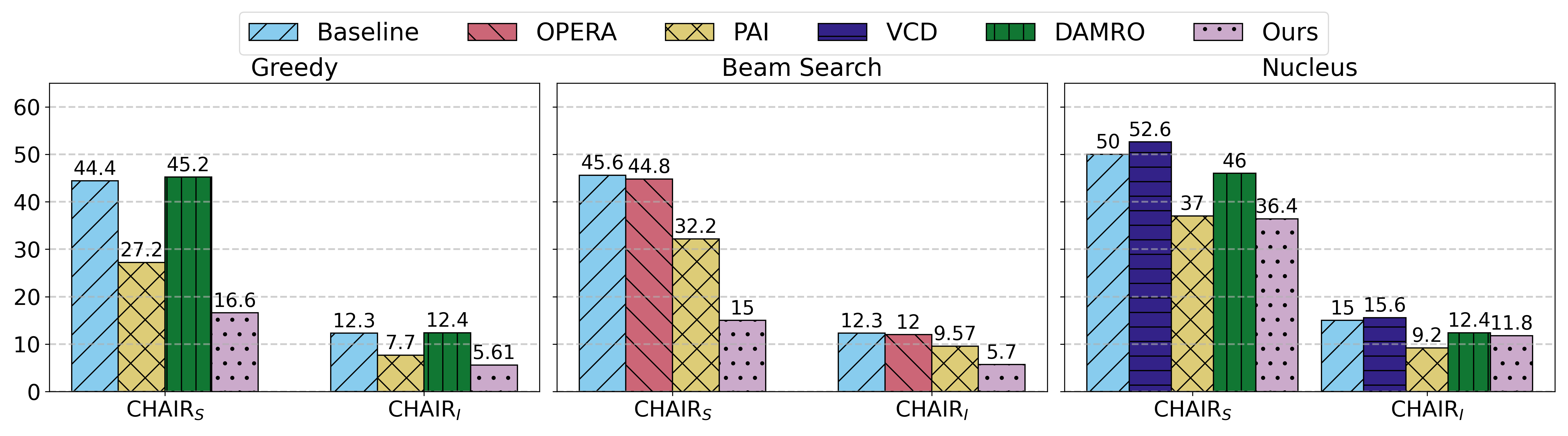}
    \includegraphics[width=0.90\linewidth]{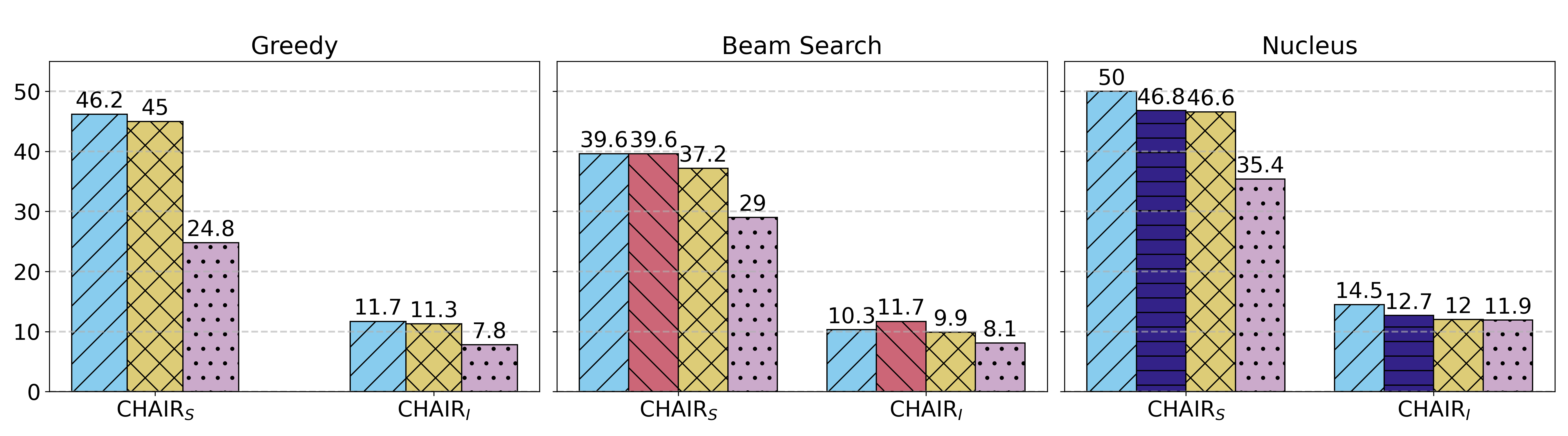}
    \vspace{-3mm}
    \caption{CHAIR scores for SPIN compared with existing approaches for greedy, beam-search, and nucleus sampling based decoding on LLaVA-1.5 (7B) (top) and Qwen-VL (bottom).}
    \label{fig:chair_decoding}
\end{figure*}

\subsection{Experimental Setup}
We evaluate our approach on LLaVA-1.5 (7B, 13B) \cite{Liu_2024_llava1.5}, LLaVA-Next \cite{liu2024llavanext}, MiniGPT-4 \cite{zhu2024minigpt}, Shikra \cite{chen2023shikra}, Qwen-VL \cite{Qwen-VL}, and Qwen2.5-VL \cite{Qwen2.5-VL} models using POPE~\cite{li2023pope}, CHAIR~\cite{CHAIR}, MMHal-Bench \cite{sun-etal-2024-mmhalbench}, MME \cite{fu2023mme}, MMMU \cite{yue2023mmmu}, and GPT-4o assisted evaluation. While POPE and CHAIR evaluate object hallucinations, with POPE using Yes-No questions and CHAIR assessing image descriptions, MMHal-Bench takes into account logical considerations like object count, attributes, and relationships. We report results using different decoding strategies including greedy, nucleus sampling, and beam-search (with beam width of 5), and compare our approach with existing methods like OPERA, VCD, PAI, and DAMRO for each decoding strategy. Notably, OPERA is only compatible with beam-search, while VCD is specifically designed for nucleus sampling. We report DAMRO results only for LLaVA models, as its applicability to more complex projection modules like Q-Former remains unclear, a limitation also acknowledged in the original paper. The hyperparameters used for the existing methods for each model are taken from their respective papers (further details are provided in the Appendix \ref{sec:baseline_hyperparams}). Additionally, to evaluate the efficiency of our method, we report throughput measured in tokens per second.  
All experiments are run in Pytorch using  NVIDIA RTX A6000 GPUs.  


\subsection{CHAIR Evaluation}

\begin{table}[t!]
\small\addtolength{\tabcolsep}{-3.5pt}
    \centering
    \resizebox{\columnwidth}{!}{
    \begin{tabular}{c|c|c|c|c|ccc}
    \toprule
    \textbf{Model} & \textbf{Method} & \textbf{Layers} & \textbf{$r$} & \textbf{$\alpha$} & \textbf{C$_S$} & \textbf{C$_I$} & \textbf{F1} \\
    \midrule
     \multirow{5}{*}{\makecell{LLaVA- \\1.5 (7B)}} & Baseline & - & - & - & 44.4 & 12.3 & \textbf{77.8}  \\
      & PAI & - & - & - & 27.2 & 7.7 & 76.8 \\
      & DAMRO & - & - & - & 45.2 & 12.4 & \textbf{77.8}\\
      & VTI & - & - & - & 35.8 & 11.1& 76.8 \\
& \lvio SPIN & \lvio 1$\sim$32 & \lvio 0.05 & \lvio 0.08 & \lvio 26.4 & \lvio 7.6 & \lvio 77.6 \\
& \lvio SPIN & \lvio 1$\sim$32 & \lvio 0.05 & \lvio 0.01 & \lvio \textbf{16.6} & \lvio \lvio \textbf{5.6} & \lvio 74.6 \\

    \midrule
     \multirow{5}{*}{\makecell{LLaVA- \\1.5 (13B)}} & Baseline & - & - & - & 41.4 & 10.9 & 78.9\\
      & PAI & - & - & - & 37.4 & 9.2 & 79.2\\
      & DAMRO & - & - & - & 41.2 & 11.0 & 78.7\\
& \lvio SPIN & \lvio 1$\sim$16 & \lvio 0.10 & \lvio 0.0 & \lvio 30.6 & \lvio 8.3 & \lvio\textbf{79.6} \\
& \lvio SPIN & \lvio 1$\sim$20 & \lvio 0.10 & \lvio 0.0 & \lvio\textbf{29.2} & \lvio\textbf{7.9} & \lvio 79.1 \\
      \midrule
   \multirow{4}{*}{\makecell{MiniGPT-4}} &  Baseline & - & - & - & 31.4 & 11.1 & \textbf{70.6}   \\
      & PAI & - & - & - & 19.8 & 8.4 & 69.7 \\
& \lvio SPIN & \lvio 1$\sim$16 & \lvio 0.18 & \lvio 0.0 & \lvio 21.0 & \lvio \textbf{6.2} & 	\lvio 68.8 \\
& \lvio SPIN & \lvio 1$\sim$16 & \lvio 0.18 & \lvio 0.05 & \lvio \textbf{17.6}	& \lvio 8.4 & 	\lvio 68.4 \\
\midrule  
\multirow{4}{*}{\makecell{Qwen-VL}} &  Baseline & - & - & - & 46.2 & 11.7 & 76.5   \\
  & PAI & - & - & - & 45.0 & 11.3 & 76.6  \\
& \lvio SPIN & \lvio 1$\sim$20 & \lvio 0.30 & \lvio 0.08 & \lvio 29.6	& \lvio 8.5 & 	\lvio \textbf{78.0} \\
& \lvio SPIN & \lvio 1$\sim$20 & \lvio 0.30 & \lvio 0.001 & \lvio \textbf{24.8}	& \lvio \textbf{7.8} & 	\lvio 76.7 \\
\midrule
\multirow{4}{*}{\makecell{Shikra}} &  Baseline & - & - & - & 55.2	& 14.0	& 75.39
\\
& PAI & - & - & - & 34.0 & 8.5 & \textbf{76.4} \\
& \lvio SPIN & \lvio 1$\sim$32 & \lvio 0.40 & \lvio 0.0 & \lvio30.6 & \lvio8.0 & \lvio75.41\\
& \lvio SPIN & \lvio 1$\sim$32 & \lvio 0.45 & \lvio 0.0 & \lvio \textbf{24.4} & \lvio\textbf{7.1} & \lvio74.3\\
    \bottomrule
    \end{tabular}}
    \vspace{-2mm}
    \caption{CHAIR evaluation with Greedy decoding.}
    \vspace{-5mm}
    \label{tab:chair}
\end{table}

\begin{table*}[t!]
\small\addtolength{\tabcolsep}{-3pt}
\centering
\begin{tabular}{c|c|c|c|c|cccccccc}
\toprule
\textbf{Model}                 & \textbf{Mode}    & \textbf{Layers} & \textbf{Supp.} & \textbf{Scale} & 
\multicolumn{2}{c}{\textbf{Random}} & \multicolumn{2}{c}{\textbf{Popular}} & \multicolumn{2}{c}{\textbf{Adversarial}} & \multicolumn{2}{c}{\textbf{Overall}} \\
& & & \textbf{Heads} & \textbf{Factor} & Accuracy & F1 & Accuracy & F1 & Accuracy & F1 & Accuracy & F1 \\ \hline
\multirow{4}{*}{\makecell{LLaVA-1.5 \\ (7B)}} & Baseline  & - & - & - &  86.77 & 85.21  &  85.73 & 84.36 &   84.57 & 83.27 & 85.69 & 84.28\\ 
& PAI & - & - & - & 88.53  & 87.54 & 87.40 & 86.53 & 85.43 & 84.59 & 87.12 & 86.22  \\
& DAMRO & - & - & - &86.73 & 85.18 & 85.77 & 84.39 & 84.53 & 83.22 & 85.68 & 84.26 \\
& \lvio SPIN & \lvio 1$\sim$32 & \cellcolor{lviolet}0.20 & \lvio 0.1 &
 \lvio \textbf{89.47} & \lvio\textbf{88.66} & \lvio\textbf{88.53} & \lvio\textbf{87.84} & \lvio\textbf{86.63} & \lvio\textbf{85.89} & \lvio\textbf{88.21} & \lvio\textbf{87.46} \\
\midrule
\multirow{3}{*}{\makecell{LLaVA-1.5 \\ (13B)}} & Baseline & - & - & - & 86.77 & 85.81  &  86.27 & 85.56 & 82.33 & 82.07 & 85.12 & 84.48 \\ 
& PAI  & - & - & - & 87.67 & 87.03 & 86.97 & 86.60 & 83.03 & 83.14 & 85.89 & 85.59  \\
& \lvio SPIN & \lvio 1$\sim$20 & \cellcolor{lviolet} 0.35 & \lvio 0.0 & \lvio\textbf{91.17} & \lvio\textbf{90.65} & \lvio\textbf{88.83} & \lvio\textbf{88.46} & \lvio\textbf{85.53} & \lvio\textbf{85.49}
& \lvio\textbf{88.51} & \lvio\textbf{88.20} \\
\midrule

\multirow{3}{*}{\makecell{MiniGPT-4}} & Baseline & - & - & - & 83.35 & 81.28 & 75.23 & 73.79 & 75.66 & 73.52 & 78.08 & 76.20 \\
& PAI & - & - & - & 85.25 & 83.62 & \textbf{75.97} & 74.93 & \textbf{77.91} & 76.10 & \textbf{79.71} & 78.22 \\

& \lvio SPIN & \lvio 1$\sim$24 & \cellcolor{lviolet} 0.05 & \lvio 0.0 &\lvio \textbf{85.64}  & \lvio \textbf{84.56} & \lvio75.36   & \lvio \textbf{75.34}  & \lvio77.50   & \lvio \textbf{76.86} & \lvio79.50 & \lvio \textbf{78.92}     \\ 
\midrule

\multirow{3}{*}{\makecell{Shikra}} & Baseline  & - & - & - & 80.77 & 80.73 & 78.06 & 78.71
& 75.99 & 77.22 & 78.27 & 78.89 \\ 
& PAI & - & - & - & 80.73 & 80.30 & 77.90 & 78.52 & 75.90 & 77.08 & 78.18 & 78.63 \\
& \lvio SPIN & \lvio 1$\sim$16 & \lvio0.20 & \lvio 0.001 & \lvio\textbf{82.04} & \lvio\textbf{81.56} & \lvio\textbf{79.15} & \lvio\textbf{79.38} & \lvio\textbf{76.78} & \lvio\textbf{77.56} & \lvio\textbf{79.32} & \lvio\textbf{79.50} \\

\bottomrule
\end{tabular}%
\vspace{-2mm}
\caption{Multi-turn POPE evaluation across random, popular, and adversarial splits using Greedy Decoding.}
\vspace{-2mm}
\label{tab:pope}
\end{table*}

\noindent
\textbf{Method:} The Caption Hallucination Assessment with Image Relevance (CHAIR) \cite{CHAIR} is a widely used metric for evaluating object hallucinations in image captioning tasks. It features two variants: the per-instance metric C$_i$, which indicates the fraction of object instances that are hallucinated, and the per-sentence metric C$_s$, which indicates the fraction of sentences containing a hallucinated object. 

CHAIR compiles a set of ground-truth objects for each image. Any object included in the image caption that is not present in the ground-truth object set is classified as a ``hallucinated object''. We evaluate our approach on randomly sampled 500 images from the COCO 2014~\cite{lin2014coco} validation set with the prompt \textit{``Please help me describe the image in detail.''}, following the same setup as used in \citet{liu2024PAI, huang2024opera}. We report F1 scores alongside CHAIR scores to ensure that reducing hallucinations does not come at the expense of missing correct objects.


\vspace{1mm}
\noindent\textbf{Evaluation on greedy decoding:} We present CHAIR results for Baseline, PAI, DAMRO, VTI, and SPIN using Greedy decoding in Table~\ref{tab:chair}. A lower CHAIR score indicates fewer hallucinated objects and a higher F1 score indicates higher caption accuracy. SPIN reduces C$_i$ and C$_s$ by 
2.2$\times$ and 
2.7$\times$ respectively over baseline LLaVA 7B with $\sim$3\% degradation in F1 when 5\%  heads are suppressed uniformly across all layers. The drop in F1 can be mitigated using a higher scaling factor while still maintaining lower CHAIR scores compared to existing methods. For LLaVA 13B, Qwen-VL and Shikra, SPIN outperforms baseline even in terms of F1 scores while reducing C$_i$ and C$_s$ by 1.38$\times$ and 1.42$\times$ for LLaVA 13B, 1.5$\times$ and 1.9$\times$ for Qwen-VL, and 1.97$\times$ and 2.3$\times$ for Shikra. For MiniGPT-4, SPIN reduces C$_i$ and C$_s$ by 1.8$\times$ and 1.5$\times$ with 1.8\% degradation in F1. We also outperform PAI, DAMRO, and VTI for all models using greedy decoding. These significant improvements in the CHAIR scores quantify the benefits of the head pruning. We provide additional CHAIR evaluation results for LLaVA-Next and Qwen2.5-VL in the Appendix Table 12.

In general, we observe that the problematic heads lie in the first 16 to 20 layers, whereas for some models like LLaVA 7B and Shikra, they are uniformly distributed across all layers. The fraction of suppressed heads ($r$) varies significantly across models, ranging from 5\% in LLaVA 7B to 45\% in Shikra. Shikra's higher head suppression ratio likely indicates its attention heads perform more distributed or redundant functions, as opposed to LLaVA, where heads are more specialized. A detailed ablation study analyzing the impact of layer position, the fraction of suppressed heads, and the choice of scaling factor is provided in Section~\ref{sec:ablation}.

\begin{figure*}[t!]
    \centering
    \includegraphics[width=0.9\linewidth]{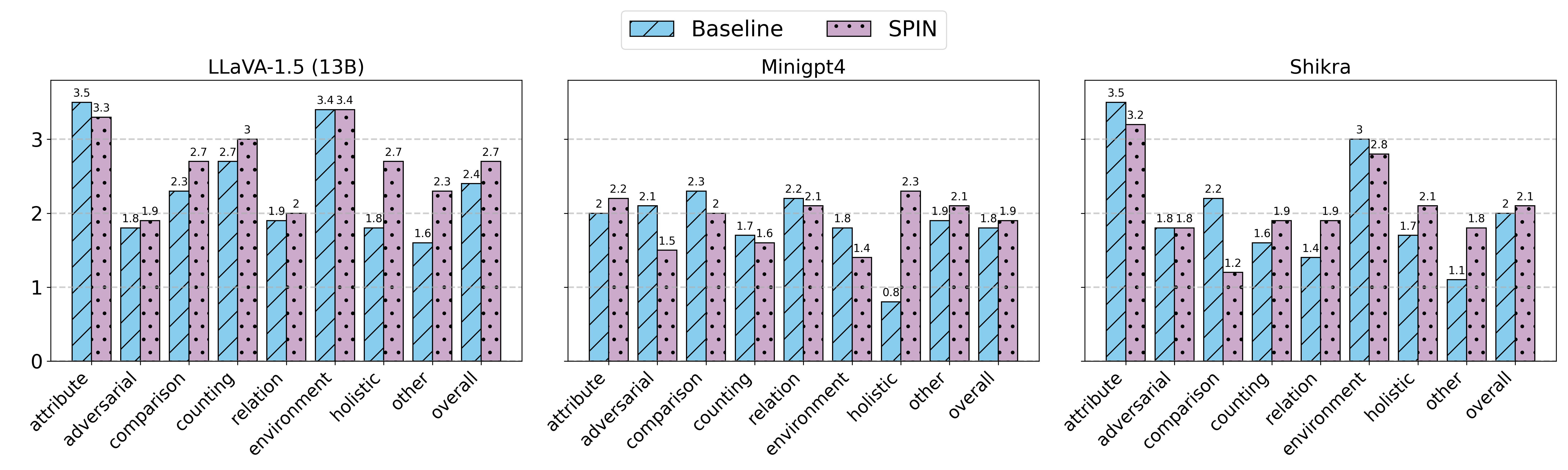}
    \vspace{-3mm}
    \caption{MMHal-Bench evaluation on LLaVA-1.5 (13B), MiniGPT-4 and Shikra.}
    \vspace{-2mm}\label{fig:mmhalbench}
\end{figure*}

\vspace{1mm}
\noindent\textbf{Evaluation on other decoding strategies:}
In Figure~\ref{fig:chair_decoding}, we compare SPIN to existing methods for different decoding modes for LLaVA 7B and Qwen-VL. We compare with OPERA for beam-search decoding, VCD for nucleus sampling, DAMRO for greedy and nucleus sampling, and PAI for all three modes. Our results demonstrate that we achieve up to 3$\times$ lower CHAIR scores compared to existing methods for greedy and beam-search. For nucleus sampling, SPIN exhibits a slight degradation in performance likely due to the variability in token selection, as further elaborated in Section \ref{sec:pope_decoding}.

\vspace{1mm}
\noindent\textbf{Mitigating Repetition from Head Pruning:}
In our experiments, we observed that SPIN occasionally leads to repetitive outputs for MiniGPT-4. While SPIN effectively reduces hallucinations and maintains F1 scores close to the baseline, it increases the length of generated sequences by up to 3.8$\times$. To mitigate this, we leverage the repetition penalty during decoding, as shown in Table~\ref{tab:rep_penalty}. While a high value can negatively impact F1, a repetition penalty of 1.1 alleviates repetitions without compromising overall output quality, as evidenced by caption length and F1 performance.

\begin{table}
\small\addtolength{\tabcolsep}{-5pt}
    \centering
    \begin{tabular}{ccccccc}
    \toprule
    \textbf{Method} & \textbf{Rep.} & \textbf{Scale} & \textbf{CHAIR$_S$} & \textbf{CHAIR$_I$} & \textbf{F1} & \textbf{Caption} \\
    & \textbf{Penalty} & \textbf{Factor} & ($\downarrow$) & ($\downarrow$) & ($\uparrow$) & \textbf{Length}\\
    \midrule
Baseline & 1.0 & - & 31.4 & 11.1 & 70.6 & 84.1 \\
    \midrule
\multirow{4}{*}{\makecell{SPIN \\ (r=0.15, \\ Layers \\ 1$\sim$24)}} & 1.0 & 0.0 & 19.6 & 6.8 & 68.2 & 321.3 \\
& 1.05 & 0.0 & 19.2 & 7.7 & 68.9 & 100.1 \\
& 1.08 & 0.0 & 20.2 & 7.8 & 68.1 & 57.4 \\
& \textbf{1.1} & \textbf{0.0} & \textbf{19.2} & \textbf{7.8} & \textbf{67.5} & \textbf{42.9} \\
\midrule
\multirow{4}{*}{\makecell{SPIN \\ (r=0.18, \\ Layers \\ 1$\sim$16)}} & 1.0 & 0.0 & 21.0 & 6.2 & 68.8 & 271.6 \\
& 1.1 & 0.0 & 17.4 & 8.3 & 67.7 & 43.0 \\
& \textbf{1.1} & \textbf{0.05} & \textbf{17.6} & \textbf{8.4} & \textbf{68.4} & \textbf{44.7} \\
& 1.1 & 0.08 & 21.6 & 9.2 & 69.4 & 46.2 \\
 \bottomrule
    \end{tabular}
    \vspace{-2mm}
    \caption{Analysis using repetition penalty on MiniGPT-4}
    \vspace{-3mm}
    \label{tab:rep_penalty}
\end{table}

\subsection{POPE Evaluation}

\noindent
\textbf{Method:} Polling-based Object Probing Evaluation (POPE) \cite{li2023pope} is used for assessing hallucinations in VQA tasks by querying models with “Is there a <object> in the image?”. The objects are drawn from three splits: \textit{random} (any dataset object), \textit{popular} (most frequent objects), and \textit{adversarial} (closely related but misleading objects). We evaluate on 500 randomly selected COCO 2014~\cite{lin2014coco} validation images, with six questions per image from each split. We present results for multi-turn POPE evaluation~\cite{liu2024PAI}, where earlier responses are appended to the input context, increasing the context length and amplifying the chances of image neglect. 

\vspace{1mm}
\noindent
\textbf{Evaluation on greedy decoding:}
Table~\ref{tab:pope} reports POPE results for LLaVA (7B and 13B), MiniGPT-4 and Shikra across all three splits using greedy decoding. SPIN consistently improves accuracy and F1 across all three splits for LLaVA and Shikra models, achieving gains of up to 3.4\% in accuracy and 3.7\% in F1 over the baseline. DAMRO performs similarly or worse than baseline for greedy decoding. SPIN consistently outperforms PAI in both accuracy and F1 for LLaVA and Shikra, achieving up to a 2.6\% gain in both metrics. For MiniGPT-4, SPIN attains the highest F1 score, while PAI achieves slightly better accuracy on the popular and adversarial splits. We provide POPE evaluation results for LLaVA-Next and Qwen2.5-VL in the Appendix Table 13.

\begin{table}[t!]
\small\addtolength{\tabcolsep}{-1pt}
\centering
\begin{tabular}{c|c|cc}
\toprule
\textbf{Decoding} & \textbf{Mode}  & \textbf{Accuracy} & \textbf{F1}  \\
\midrule
\multirow{4}{*}{\makecell{Beam \\ Search}} & Baseline & 83.58 &	81.14\\
& OPERA & 83.67 &	81.29\\
& PAI & 84.52 & 82.80 \\
& SPIN & \textbf{86.15} & \textbf{84.93}\\

\midrule
\multirow{5}{*}{\makecell{Nucleus}}  & Baseline & 82.97 &	81.70\\
& VCD & 84.42 & 83.45 \\
& PAI & 84.96 & 83.99 \\
& DAMRO & \textbf{85.69} & \textbf{84.28} \\
& SPIN & 82.49 & 81.83\\

\bottomrule
\end{tabular}
\caption{Multi-turn POPE evaluation using Beam Search and Nucleus Sampling decoding for LLaVA-1.5 (7B).}
\vspace{-1mm}
\label{tab:pope_decoding}
\vspace{-4mm}
\end{table}

\vspace{1mm}
\noindent\textbf{Evaluation on other decoding strategies:}
\label{sec:pope_decoding}
In Table \ref{tab:pope_decoding}, we report overall accuracy and F1 scores for multi-turn POPE evaluation using beam-search and nucleus sampling decoding. We observe that SPIN yields the best results for beam-search, whereas DAMRO performs the best for nucleus sampling. 
We observe that SPIN performs more effectively with greedy decoding and beam-search, but is less effective under nucleus sampling. This may be attributed to the inherent randomness in sampling-based methods, which introduce greater variability in token selection. As a result, even after suppressing hallucination-prone heads, the stochastic nature of nucleus sampling can still lead to hallucinated outputs, limiting the impact of structural interventions like head suppression.


\subsection{MMHal-Bench Evaluation}

\noindent
\textbf{Method:} To extend our analysis to more complex and logically challenging benchmarks, we evaluate using MMHal-Bench \cite{sun-etal-2024-mmhalbench}. This benchmark comprises 96 carefully curated image-question pairs based on images from OpenImages \cite{openimages}, each accompanied by ground-truth answers. The pairs span 12 common object meta-categories derived from COCO, and the questions test nuanced reasoning across eight categories: object attributes, adversarial objects, comparisons, counting, spatial relations, environment, holistic descriptions, and others—which includes cases where models fail to recognize text, misinterpret icons, or incorrectly reason about the observed visual content. We generate responses using both the baseline model and SPIN, and then employ GPT-4 to score each response based on its agreement with the ground-truth answers. 

\vspace{1mm}
\noindent
\textbf{Evaluation Results:} The final evaluation report including per-category scores for each model, with the overall performance measured as the average across all categories, is presented in Figure \ref{fig:mmhalbench}. We observe that SPIN improves overall scores over the baseline for all three models: LLaVA-1.5, MiniGPT-4 and Shikra, with the maximum improvements observed in \textit{holistic description} and \textit{other} categories. Note, we use the same values for Layers, $r$, and $\alpha$ as reported for CHAIR evaluation.

\begin{table*}[t!]
\small\addtolength{\tabcolsep}{-3pt}
\centering
\resizebox{\textwidth}{!}{
\begin{tabular}{c|c|ccccccccccc|ccccc}
\toprule    

\multirow{3}{*}{\textbf{Model}} & \multirow{3}{*}{\textbf{Method}} & \multicolumn{11}{c}{\textbf{Perception}} & \multicolumn{5}{c}{\textbf{Cognition}} \\
& & \makecell{Exis-\\tence} & Count & \makecell{Posi-\\tion} & Color & Posters & \makecell{Cele-\\brity} & Scene & \makecell{Land-\\mark} & \makecell{Art-\\work} & OCR & Total & \makecell{Comm-\\onsense} & \makecell{Num-\\erical} & Text & Code & Total \\

\midrule
\multirow{2}{*}{\makecell{LLaVA-\\ 1.5 (7B)}} & Baseline & 190.0 & 108.3 & 96.7 & 135.0 & 140.8 & 134.7 & 153.5 & 140.0 & 75.3 & 92.5 & 1266.8 & 93.6 & 77.5 & 45.0 & 45.0 & 261.1 \\
& SPIN & 190.0 & \textbf{113.3} & \textbf{101.7} & 135.0 & \textbf{142.9} & \textbf{137.4} & 151.8 & 138.5 & \textbf{76.0} & \textbf{107.5} & \textbf{1294.0} & 93.6 & 75.0 & \textbf{60.0} & 45.0 & \textbf{273.6} \\
\midrule
\multirow{2}{*}{\makecell{LLaVA-\\ 1.5 (13B)}} & Baseline & 185.0 & 110.0 & 100.0 & 135.0 & 146.6 & 121.8 & 161.3 & 160.3 & 87.0 & 117.5 & 1324.4 & 110.7 & 52.5 & 90.0 & 47.5 & 300.7 \\
& SPIN & 185.0 & 110.0 & 100.0 & 135.0 & 144.2 & 121.5 & \textbf{162.0} & 158.8 & \textbf{90.8} & 117.5 & \textbf{1324.7} & \textbf{112.1} & \textbf{62.5} & 87.5 & 47.5 & \textbf{309.6} \\
\midrule
\multirow{2}{*}{\makecell{Qwen2.5-\\ VL (7B)}} & Baseline & 180.0 & 120.0 & 148.3 & 190.0 & 169.4 & 117.9 & 139.5 & 123.3 & 141.0 & 192.5 & 1521.9 & 110.7 & 155.0 & 185.0 & 140.0 & 590.7 \\
& SPIN & \textbf{185.0} & \textbf{125.0} & 148.3 & 190.0 & \textbf{173.1} & \textbf{135.6} & 137.5 & \textbf{131.3} & 134.0 & 177.5 & \textbf{1537.3} & 110.0 & 125.0 & 185.0 & 132.5 &552.5 \\
\bottomrule
\end{tabular}}
\vspace{-2mm}
\caption{Results on MME evaluation. SPIN improves performance across both perception and cognition subsets on all three models. A small degradation in cognitive performance is observed for Qwen2.5-VL.}
\label{tab:mme}
\end{table*}

\begin{table*}[t!]
\small\addtolength{\tabcolsep}{-2.5pt}
\centering
\begin{tabular}{c|c|cccccc|c}
\toprule
\textbf{Model} & \textbf{Mode} & \textbf{\makecell{Art \& \\Design}} & \textbf{Business} & \textbf{Science} & \textbf{\makecell{Health \&\\ Medicine}} & \textbf{\makecell{Human. \& \\Social Sci.}} & \textbf{\makecell{Tech \& \\Eng.}} & \textbf{Overall} \\

\midrule
\multirow{2}{*}{\makecell{LLaVA-1.5 (7B)}} & Baseline & 51.7 & 23.3 & 24.7 & 34.0 & 49.2 & 31.4 & 34.4 \\
& SPIN & 50.0 & \textbf{24.0} & \textbf{25.3} & \textbf{35.3} & 47.5 & 31.4 & 34.4 \\
\midrule
\multirow{2}{*}{\makecell{LLaVA-1.5 (13B)}} & Baseline & 51.7 & 23.3 & 28.0 & 39.3 & 52.5 & 32.4 & 36.6 \\
& SPIN & 51.7 & 23.3 & \textbf{29.3} & 38.0 & \textbf{54.2} & \textbf{32.9} & \textbf{36.9} \\
\midrule
\multirow{2}{*}{\makecell{LLaVA-Next (7B)}} & Baseline & 52.5 & 30.0 & 20.7 & 36.7 & 57.5 & 26.2 & 35.3 \\
& SPIN & 52.5 & 29.3 & 20.7 & 34.0 & 57.5 & \textbf{29.5} & \textbf{35.6} \\
\midrule
\multirow{2}{*}{\makecell{Qwen2.5-VL (7B)}} & Baseline & 68.3 & 40.0 & 40.0 & 55.3 & 67.5 & 39.5 & 49.9 \\
& SPIN & 67.5 & \textbf{42.7} & 37.3 & 54.7 & \textbf{68.3} & \textbf{40.0} & 49.9 \\

\bottomrule

\end{tabular}
\vspace{-1mm}
\caption{Performance on MMMU validation set for evaluating general capability. SPIN maintains or slightly improves general multimodal reasoning capabilities across all evaluated LVLMs. Following the original MMMU \cite{yue2023mmmu} evaluation protocol, the overall score is computed as a weighted average across all subtasks.}
\vspace{-2mm}
\label{tab:mmmu}
\end{table*}

\subsection{GPT-4o Assisted Evaluation}

\begin{table}[!t]
\small\addtolength{\tabcolsep}{-3pt}
\centering
    \begin{tabular}{ccccccc}
    \toprule
    \textbf{Method} & \multicolumn{2}{c}{\textbf{LLaVA-1.5 (7B)}} & \multicolumn{2}{c}{\textbf{LLaVA-1.5 (13B)}} & \multicolumn{2}{c}{\textbf{Qwen-VL}} \\
    & A & D & A & D & A & D \\
    \midrule
    Baseline & 6.50 & 6.83 & 6.53 & 6.71 & 6.56 & 6.87\\
    SPIN & 7.19 & 6.12 & 7.03 & 6.18 & 7.08 & 6.22\\
    \bottomrule
    \end{tabular}
    \caption{GPT-4o assisted evaluation showcasing Accuracy (A) and Detailedness (D) of the generated captions.}
    \label{tab:GPT-4o_eval}
    \vspace{-4mm}
\end{table}

We use GPT-4o to compare the responses of baseline and SPIN models using Greedy decoding on a scale of 1 to 10 using two metrics: (1) \textit{Accuracy (A):} measuring if the description is accurate with respect to the image content, and (2) \textit{Detailedness (D):} measuring the richness of necessary details in the responses. The prompt we used is given in the Appendix Table 14.
It is designed to ensure that the sequential order in which the responses are presented does not affect the judgement. The evaluation model is prompted to identify objects misaligned with the image context, identifying any discrepancies in count, position, and colors of objects in the images. The results presented in Table \ref{tab:GPT-4o_eval} show that SPIN improves Accuracy up to $\sim$7\%, while maintaining Detailedness close to baseline. We use the same values for Layers, $r$, and $\alpha$ reported for CHAIR for all captioning tasks.

\subsection{General Performance Assessment}
\noindent\textbf{MME Evaluation:} In Table \ref{tab:mme}, we present SPIN results on MME full set \cite{fu2023mme} that assesses the perceptual and cognitive abilities of VLMs across a total of 14 subtasks, including tasks such as OCR, visual knowledge, attribute relationships, and object recognition. SPIN outperforms the baseline on both perception and cognition tasks for all three models, with a small degradation in cognition for Qwen2.5-VL. The improvements in the first four subtasks relating to existence, count, position, and colour demonstrates the model’s ability to reduce object-level and attribute-level hallucinations.

\vspace{1mm}
\noindent\textbf{MMMU Evaluation:} In Table \ref{tab:mmmu}, we present MMMU \cite{yue2023mmmu} (a standard multimodal understanding and reasoning benchmark) scores for SPIN to assess its impact on the general capability of LVLMs. SPIN attains the same performance as baseline for LLaVA-1.5 (7B) and Qwen2.5-VL, while attaining a small improvement on LLaVA-1.5 (13B) and LLaVA-Next. This demonstrates that head suppression does not hinder, but enhances the model's core capabilities while providing substantial hallucination reduction. Notably, we use the same parameters for head suppression (Layers, $r$, and $\alpha$) as reported for CHAIR evaluation for both MME and MMMU.

\begin{table*}[t]
\small\addtolength{\tabcolsep}{-1pt}
  \begin{minipage}{0.33\textwidth}
      \caption*{LLaVA-1.5 (7B)}
      \vspace{-2mm}
    \centering
        \begin{tabular}{ccccc}
    \toprule
    \textbf{Layers} & \textbf{$r$} & \textbf{C$_S$} & \textbf{C$_I$} & \textbf{F1} \\
    &  & ($\downarrow$) & ($\downarrow$) & ($\uparrow$) \\
    \midrule
    \lblue 1$\sim$32 & \lblue 0.05 & \lblue \textbf{17.4} & \lblue \textbf{5.8} & \lblue 74.6 \\
    1$\sim$32 & 0.10 & 22.8 & 7.4 & 74.1 \\
    1$\sim$32 & 0.15 & 20.8 & 6.4 & 73.8 \\
    \midrule
    1$\sim$16 & 0.05 & 18.2 & 6.2 & 73.7 \\
 1$\sim$20 & 0.05 & 18.4 & 6.3 & 74.1 \\
 1$\sim$24 & 0.05 & 19.2 & 6.5 & 73.8 \\
 4$\sim$32 & 0.05 & 42.0 & 11.4 & 78.2 \\
 8$\sim$32 & 0.05 & 42.4 & 11.1 & \textbf{78.8} \\
\bottomrule
    \end{tabular}
  \end{minipage}
  \begin{minipage}{0.30\textwidth}
      \caption*{LLaVA-1.5 (13B)}
      \vspace{-2mm}    
      \centering
        \begin{tabular}{ccccc}
    \toprule
    \textbf{Layers} & \textbf{$r$} & \textbf{C$_S$} & \textbf{C$_I$} & \textbf{F1} \\
    &  & ($\downarrow$) & ($\downarrow$) & ($\uparrow$) \\
    \midrule
    1$\sim$32 & 0.05 & 45.8 & 12.0 & 78.2 \\
\lblue1$\sim$32 & \lblue0.10 & \lblue\lblue29.6 & \lblue8.0 & \lblue78.9 \\
1$\sim$32 & 0.15 & 41.6 & 10.3 & \textbf{80.2} \\
1$\sim$32 & 0.20 & 36.0 & 10.0 & 79.1 \\
\midrule
\lylw 1$\sim$16 & \lylw 0.10 & \lylw 30.6 & \lylw 8.3 & \lylw 79.6 \\
\lylw 1$\sim$20 & \lylw 0.10 & \lylw 29.2 & \lylw \textbf{7.9} & \lylw 79.1 \\
1$\sim$24 & 0.10 & \textbf{28.0} & 9.2 & 79.3 \\
\bottomrule
    \end{tabular}
  \end{minipage}
  \begin{minipage}{0.30\textwidth}
      \caption*{Qwen-VL}
      \vspace{-3mm}
    \centering
        \begin{tabular}{cccccc}
    \toprule
    \textbf{Layers} & \textbf{$r$} & \textbf{C$_S$} & \textbf{C$_I$} & \textbf{F1} \\
    &  & ($\downarrow$) & ($\downarrow$) & ($\uparrow$) \\
    \midrule
1$\sim$32 & 0.05 & 44.4 & 11.7 & \textbf{77.6} \\
1$\sim$32 & 0.10 & 39.8 & 10.9 & 77.1 \\
1$\sim$32 & 0.15 & 41.6 & 11.7 & 77.5 \\
1$\sim$32 & 0.20 & 39.6 & 11.2 & 76.9 \\
1$\sim$32 & 0.25 & 36.4 & 9.7 & 77.3 \\
\lblue1$\sim$32 & \lblue0.30 & \lblue\textbf{26.0} & \lblue9.1 & \lblue75.6 \\
\midrule
\lylw1$\sim$20 & \lylw0.30 & \lylw27.6 & \lylw\textbf{7.9} & \lylw76.1 \\
1$\sim$24 & 0.30 & 28.4 & 8.3 & 76.2 \\
8$\sim$32 & 0.30 & 32.4 & 9.4 & 76.3
\\
\bottomrule
    \end{tabular}
  \end{minipage}
  \vspace{-2mm}
  \caption{Ablation on the ratio of suppressed heads and layer selection for three different models. The best $r$ is identified in the top part of the tables (marked with \colorbox{blue!15}{blue}). The layer configuration for selected value of $r$ is presented in the bottom part (best layer configuration is marked in \colorbox{yellow!40}{yellow}).}
  \label{tab:ablation_head}
\end{table*}

\begin{figure*}[t!]
    \centering
    \includegraphics[width=0.99\linewidth]{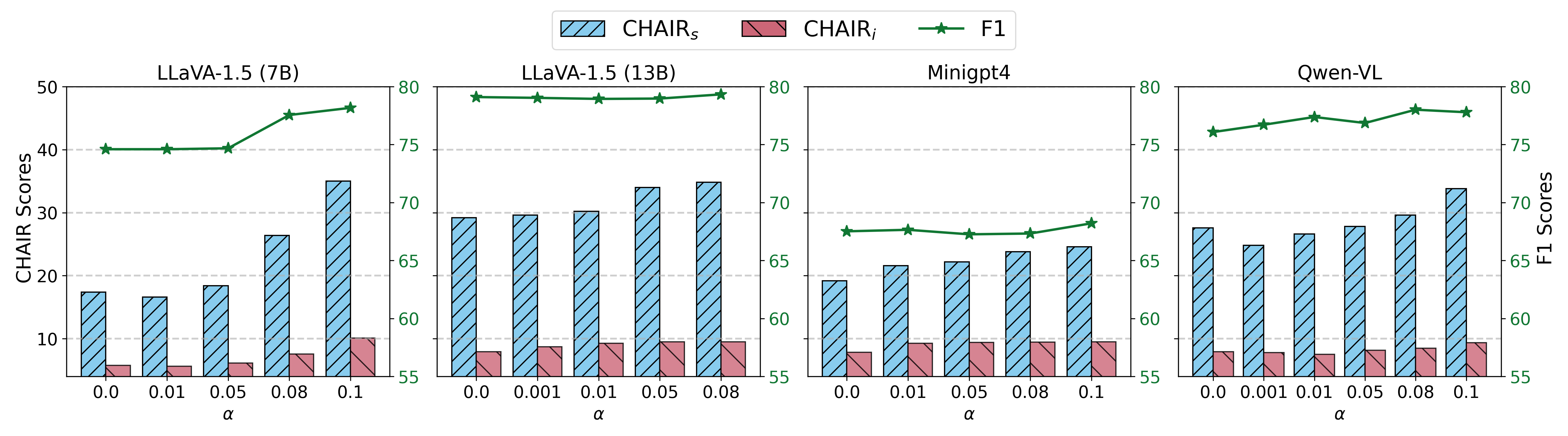}
    \vspace{-4mm}
    \caption{Ablation on the suppression factor ($\alpha$)  for four models.}
    \vspace{-2mm}
    \label{fig:ablation_alpha}
\end{figure*}

\subsection{Throughput Estimation}
\begin{figure}[htbp]
    \centering
    \vspace{-3mm}
    \includegraphics[width=0.84\linewidth]{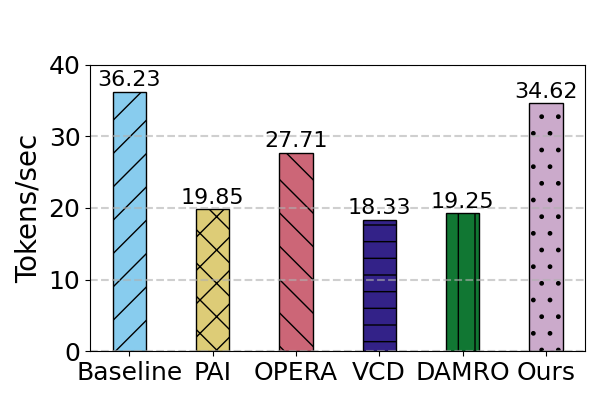}
    \vspace{-2mm}
    \caption{Throughput comparison with existing methods given by the number of tokens generated per second.}
    \label{fig:tokens_per_sec}
\end{figure}

To assess the efficiency of different hallucination mitigation strategies, we compare their throughput in Figure \ref{fig:tokens_per_sec}. Throughput is computed by measuring the total computational latency 
for generating tokens, and dividing the total number of generated tokens by this latency. SPIN achieves the highest throughput among SoTA approaches, while achieving throughput performance close to baseline.
\vspace{-2mm}

\subsection{Ablation Studies}
We present the hyperparameter evaluations for SPIN in Table \ref{tab:ablation_head}, where we first vary $r$ to obtain reduced hallucinations without a significant drop in F1. We obtain $r=0.05$ for LLaVA 7B, 0.10 for LLaVA 13B, and 0.30 for Qwen-VL. Next, we evaluate different layer configurations to find hallucination-prone layers
\label{sec:ablation}. While uniformly pruning across all layers works best for LLaVA 7B, initial layers are more hallucination-prone for LLaVA 13B and Qwen-VL. Finally, for selected $r$ and layers for head suppression, we plot $C_s$, $C_i$ and F1 for different $\alpha$ in Figure \ref{fig:ablation_alpha}. F1 scores show comparable values for different $\alpha$ for MiniGPT-4 and LLaVA 13B models. However, for LLaVA 7B and Qwen-VL, complete pruning leads to a drop in F1. 
This happens because the pre-trained model weights are not adapted to this change. Therefore, we perform partial head suppression using the parameter $\alpha$, where a higher value of $\alpha$ preserves the F1 score by reducing the suppression intensity.



\section{Conclusions}
Despite achieving remarkable progress in vision-language tasks, LVLMs still suffer from hallucinations arising from misalignment between visual input and the generated text. We identify a dynamic set of attention heads based on input query as the potential cause for hallucinations. To counteract this, we propose an image-attention guided head pruning strategy agnostic to decoding method, that is run-time efficient and can be seamlessly integrated during inference. Our approach leads to a substantial reduction in hallucinations, outperforming existing methods while maintaining accuracy.

\section*{Limitations}
We propose an inference-only solution for ease of integration and to accommodate scenarios with limited training resources. However, our approach could be enhanced by incorporating a trainable router for more adaptive head selection. Our method's effectiveness is somewhat reduced when paired with sampling-based decoding strategies (e.g., nucleus sampling). We attribute this to the greater token selection variability inherent in these stochastic approaches (as further detailed in Section \ref{sec:pope_decoding}).
Our method requires access to the model weights, which limits its applicability to open-source models. It cannot be applied to closed-source or API-based models, where direct access to weights is restricted.


\paragraph{Usage and License:}
We acknowledge the licensing terms associated with the artifacts used in this study. LLaVA~\cite{Liu_2024_llava1.5} is released under an open-source license (Apache 2.0), permitting modification and distribution with attribution. MiniGPT-4~\cite{zhu2024minigpt} follows similar licensing terms under the LLaMA model’s restrictions, where the base model weights are subject to Meta’s research agreement. Qwen-VL models are also released under Apache 2.0, whereas Shikra is released under the Creative Commons Attribution-NonCommercial 4.0 International (CC BY-NC 4.0) license. Our use of these models and datasets is strictly for academic research purposes, adhering to all licensing and usage guidelines. All datasets used in our experiments comply with their respective licenses, and we ensure proper attribution where required.

\bibliography{custom}

\appendix
\input{appendix}

\end{document}

%% file: appendix.tex
\clearpage
\section{Appendix}
\label{sec:appendix}

\subsection{Baseline Hyperparameters}
\label{sec:baseline_hyperparams}
The hyperparameters \( \alpha \) and \( r \) for SPIN are specified in each table. For existing methods, we use the best-performing configurations reported in their respective papers.

For PAI, we adopt the following hyperparameter settings: for LLaVA, we use \( \alpha = 0.5 \), \( \gamma = 1.2 \), and \(layer\ prior=2\); for MiniGPT-4, \( \alpha = 0.2 \), \( \gamma = 1.3 \), and \(layer\ prior=3\); and for Shikra, \( \alpha = 0.6 \), \( \gamma = 1.1 \), and \(layer\ prior=3\). Since the original PAI paper does not report results on Qwen-VL, we experimented with several configurations and found that reusing Shikra's settings yielded the best performance.

For OPERA, we use the same configuration across all models: \( \sigma = 50 \), \( r = 15 \), \( N_{\text{can}} = 5 \), and \( \alpha = 1 \).  
For VCD, we use \( \alpha = 1 \), \( \beta = 0.1 \), and apply noise steps \( T = 999 \) for all models. For DAMRO, we select the top 10 tokens as outliers and set \( \alpha = 0.5 \).

\subsection{Ablation Studies}

\noindent\textbf{Head Selection Strategy:}
We explore a variety of training-free head selection strategies, as summarized in Table~\ref{tab:ablation_head_strategy}. Query-based head selection proposed in \cite{jin2024moh} is computed based on the L2 norm of the query tokens, whereas the key based selection is performed based on the L2 norm of the key tokens. In addition, we experiment with head selection guided by the overall attention (to both text and image tokens) as well as attention to the image tokens only. Among these strategies, selecting heads based solely on image attention proves to be the most effective for hallucination reduction, yielding the best CHAIR scores. Here, we use \(\alpha = 0\), completely pruning attention heads across all layers.  

\begin{table}[h]
\small\addtolength{\tabcolsep}{-3pt}
    \centering
    \begin{tabular}{c|c|ccc}
    \toprule
    \textbf{Selection} & \textbf{Supp.} & \textbf{CHAIR$_S$} & \textbf{CHAIR$_I$} & \textbf{F1} \\
    \textbf{Strategy} & \textbf{Heads} & ($\downarrow$) & ($\downarrow$) & ($\uparrow$) \\
    \midrule
    Query & 0.10 & 40.4 & 11.6 & 77.4\\
    Query & 0.25 & 34.6 & 10.4 & 75.8\\
    Key & 0.25 & 44.6 & 12.4 & 77.1\\
    Attention & 0.15 & 20.4 & 8.0 & 73.4 \\
    Attention & 0.05 & 17.6 & 6.1 & 74.5 \\
    Image Attention & 0.05 & \textbf{17.4} & \textbf{5.8} & 74.6\\
    \bottomrule
    \end{tabular}
    \vspace{-2mm}
    \caption{Comparison of different training-free head selection strategies in SPIN on LLaVA-1.5 (7B) using CHAIR evaluation. Selecting heads based on image attention yields the most effective reduction in hallucination.}
    \vspace{-7mm}
    \label{tab:ablation_head_strategy}
\end{table}

\subsection{Head Mask Visualization} 
\begin{figure}[htbp]
    \centering
    \includegraphics[trim = 10 20 20 30, clip, width=0.9\linewidth]{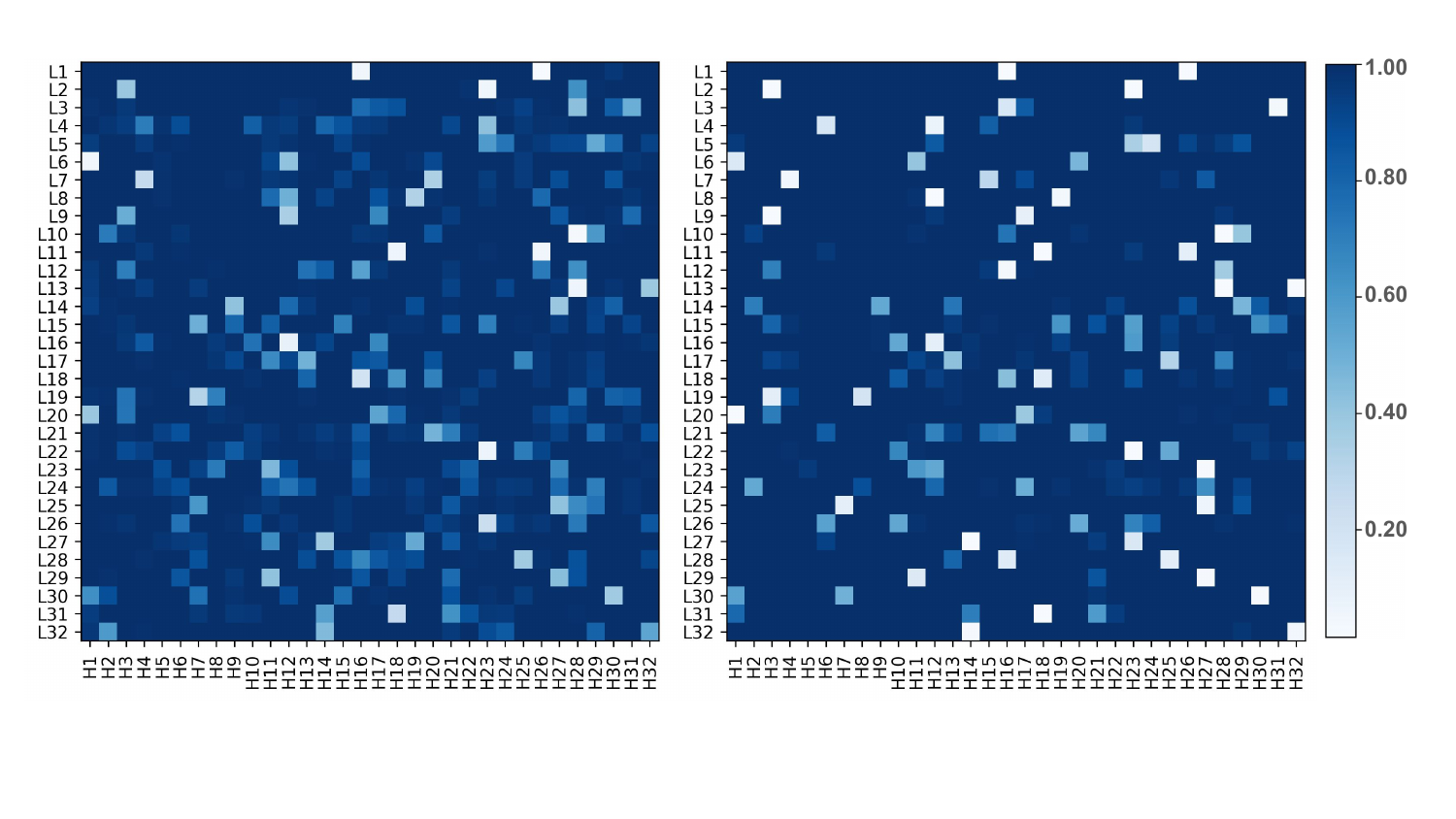}
    \vspace{-7mm}    
    \caption{Heatmap for head masks averaged over all test samples for CHAIR (left) and POPE random split (right) for 95\% active heads. White and dark blue indicate the always pruned and always active heads, respectively.}
    \vspace{-6mm}
    \label{fig:heatmap}
\end{figure}

The head masks for POPE and CHAIR using 95\% active heads
are visualized in Figure~\ref{fig:heatmap}. The pruned heads are more consistent for POPE, likely due to the fact that ``Yes'' or ``No'' answers are generated as the first token. In contrast, CHAIR generates longer sequences, causing tokens at each position to likely prune different heads. 

\begin{table}
\small\addtolength{\tabcolsep}{-5pt}
\centering
\begin{tabular}{c|c|ccc|cc}
\toprule
\textbf{Model} & \textbf{Mode} & \textbf{Layers} & \textbf{$r$} & \textbf{$\alpha$} & \textbf{Accuracy} & \textbf{F1} \\

\midrule

\multirow{4}{*}{\makecell{LLaVA-1.5 \\(7B)}} 
& Baseline & - & - & - & 85.69 & 84.28 \\
& SPIN & 1$\sim$16 & 0.15 & 0.0 & 87.84 & 87.13 \\
& SPIN & 1$\sim$16 & 0.15 & 0.001 & 87.79 & 87.07 \\
& SPIN & 1$\sim$32 & 0.20 & 0.08 & \textbf{88.19} & \textbf{87.42} \\

\midrule

\multirow{4}{*}{\makecell{LLaVA-1.5 \\(13B)}} 
& Baseline & - & - & - & 85.12 & 84.48  \\
& SPIN & 1$\sim$32 & 0.35 & 0.0 & 88.47 & 88.00  \\
& SPIN & 1$\sim$24 & 0.35 & 0.0 & 88.40 & 88.01  \\
& SPIN & 1$\sim$20 & 0.35 & 0.001 & \textbf{88.50} & \textbf{88.17}  \\

\midrule

\multirow{4}{*}{\makecell{MiniGPT-4}} 
& Baseline & - & - & - & 78.08 & 76.20  \\
& SPIN & 1$\sim$32 & 0.05 & 0.0 & \textbf{79.43} & \textbf{78.78}  \\
& SPIN & 1$\sim$24 & 0.05 & 0.08 & 79.37 & 78.61  \\
& SPIN & 1$\sim$24 & 0.05 & 0.001 & 79.14 & 78.46  \\

\midrule

\multirow{4}{*}{\makecell{Shikra}} 
& Baseline & - & - & - & 78.27 & 78.89  \\
& SPIN & 1$\sim$16 & 0.20 & 0.0 & \textbf{79.20} & 79.40 \\
& SPIN & 1$\sim$16 & 0.20 & 0.01 & 79.17 & 79.37 \\
& SPIN & 1$\sim$16 & 0.20 & 0.05 & 79.19 & \textbf{79.45} \\

\bottomrule
\end{tabular}
\vspace{-2mm}
\caption{Multi-turn POPE evaluation under greedy decoding. SPIN consistently reduces hallucinations across all models on multi-turn vision-language tasks.}
\vspace{-5mm}
\label{tab:more_pope_results}
\end{table}

\subsection{Extended Evaluation of SPIN}



\begin{table*}[bthp]
\small\addtolength{\tabcolsep}{-1pt}
    \centering
    \begin{tabular}{c|c|c|ccc||ccc}
    \toprule
    \textbf{Model} & \textbf{Decoding} & \textbf{Method} & \textbf{Layers} & \textbf{Supp.} & \textbf{Scale} & \textbf{CHAIR$_S$} & \textbf{CHAIR$_I$} & \textbf{F1} \\
    & & & & \textbf{Heads} & \textbf{Factor} & ($\downarrow$) & ($\downarrow$) & ($\uparrow$) \\
\midrule



\multirow{9}{*}{\makecell{MiniGPT-4}} 


& \multirow{5}{*}{\makecell{Beam \\Search}} 
& Baseline & - & - & - & 33.4 & 11.0 & \textbf{70.2} \\
& & OPERA & - & - & - & 23.2 & 8.8 & 69.4 \\
& & PAI & - & - & - & 22.2 & 8.3 & 68.1 \\
& & SPIN & 1$\sim$24 & 0.15 & 0.0 & 16.6 & \textbf{4.8} & 68.2 \\
& & SPIN & 1$\sim$16 & 0.18 & 0.05 & \textbf{16.2} & 6.6 & 69.1 \\

\cmidrule{2-9}

& \multirow{3}{*}{\makecell{Nucleus \\Sampling}} 
& Baseline & - & - & - & 31.8 & 11.8 & 66.4 \\
& & PAI & - & - & - & \textbf{23.0} & \textbf{9.0} & \textbf{68.3} \\
& & SPIN & 1$\sim$16 & 0.18 & 0.05 & 23.6 & 12.5 & 62.9 \\

\midrule



\multirow{9}{*}{\makecell{Shikra}} 


& \multirow{4}{*}{\makecell{Beam \\Search}} 
& Baseline & - & - & - & 52.6 & 14.1 & \textbf{75.7} \\
& & OPERA & - & - & - & 40.2 & 12.6 & 72.7 \\
& & PAI & - & - & - & 38.6 & 10.1 & 74.9 \\
& & SPIN & 1$\sim$32 & 0.40 & 0.0 & \textbf{32.4} & \textbf{10.0} & 74.5 \\

\cmidrule{2-9}

& \multirow{4}{*}{\makecell{Nucleus \\Sampling}} 
& Baseline & - & - & - & 56.0 & 15.7 & 72.8 \\
& & VCD & - & - & - & 55.6 & 15.2 & 74.9 \\
& & PAI & - & - & - & \textbf{33.8} & \textbf{8.7} & \textbf{75.1} \\
& & SPIN & 1$\sim$32 & 0.40 & 0.0 & 34.2 & 10.6 & 71.5 \\

\bottomrule
\end{tabular}
\vspace{-2mm}
\caption{CHAIR evaluation using MiniGPT-4 and Shikra under beam search and nucleus sampling decoding.}
\vspace{-3mm}
\label{tab:chair_decoding_minigpt_shikra}
\end{table*}

\noindent\textbf{POPE Evaluation:}
We assess SPIN on POPE evaluation to examine its performance in multi-turn dialogue settings. All experiments in this section are performed using greedy decoding. As shown in Table~\ref{tab:more_pope_results}, SPIN consistently reduces hallucinations across multiple models. These results demonstrate that our method not only generalizes well beyond CHAIR, but also achieves strong hallucination mitigation on multi-turn vision-language tasks.
\vspace{1mm}

\noindent\textbf{CHAIR Results under Diverse Decoding Strategies:}
To further evaluate the effectiveness of SPIN, we conduct experiments on CHAIR under various decoding strategies: greedy search, beam search, and nucleus sampling. Across four models: LLaVA-1.5 (7B), MiniGPT-4, Shikra, and Qwen-VL. We explore a wider range of configurations by varying the ratio of suppressed heads, scaling factors, and the layers where SPIN is applied.

As discussed in Section~4.2, we have already provided a detailed analysis of SPIN's performance under greedy search for all four models, as well as its behavior on beam search and nucleus sampling for LLaVA-1.5 (7B) and Qwen-VL. Here, we highlight the results for MiniGPT-4 and Shikra under beam search and nucleus sampling, as shown in Table~\ref{tab:chair_decoding_minigpt_shikra}. For MiniGPT-4, SPIN reduces CHAIR$_\text{I}$ and CHAIR$_\text{S}$ by 2.3$\times$ and 2.0$\times$ with beam search. For Shikra, the corresponding reductions are 1.4$\times$ and 1.6$\times$. Under nucleus sampling, SPIN yields less significant improvements, but still effectively mitigates hallucinations. For MiniGPT-4, SPIN reduces CHAIR$_\text{S}$ by 1.35$\times$. On Shikra, it reduces CHAIR$_\text{I}$ and CHAIR$_\text{S}$ by 1.5$\times$ and 1.6$\times$, with only a 1.3\% drop in F1.

\begin{table}[t!]
\small\addtolength{\tabcolsep}{-3.5pt}
    \centering
    \begin{tabular}{c|c|ccc|ccc}
    \toprule
    \textbf{Model} & \textbf{Method} & \textbf{Layers} & \textbf{$r$} & \textbf{$\alpha$} & \textbf{C$_S$($\downarrow$)} & \textbf{C$_I$($\downarrow$)} & \textbf{F1($\uparrow$)} \\
    \midrule
     \multirow{3}{*}{\makecell{LLaVA- \\Next (7B)}} & Baseline & - & - & - & 35.6 & 8.6 & 71.7 \\
& SPIN & 1$\sim$32 & 0.05 & 0.0 & 32.6 & 8.1 & \textbf{72.3} \\
& SPIN & 1$\sim$16 & 0.20 & 0.05 & \textbf{26.8} & \textbf{7.4} & 71.7 \\

    \midrule
     \multirow{3}{*}{\makecell{Qwen2.5- \\VL (7B)}} & Baseline & - & - & - & 34.8 & 7.9 & \textbf{75.7} \\
& SPIN & 1$\sim$16 & 0.02 & 0.0 & \textbf{28.6} & \textbf{7.0} & 74.0 \\
& SPIN & 1$\sim$16 & 0.02 & 0.05 & 30.6 & 7.5 & 74.7 \\

\bottomrule
\end{tabular}
\vspace{-2mm}
\caption{CHAIR evaluation results on the latest LVLMs, including LLaVA-Next and Qwen2.5-VL.}
\vspace{-5mm}
\label{tab:latest_models_chair}
\end{table}

\begin{table}
\small\addtolength{\tabcolsep}{-5pt}
\centering
\begin{tabular}{c|c|ccc|cc}
\toprule
\textbf{Model} & \textbf{Mode} & \textbf{Layers} & \textbf{$r$} & \textbf{$\alpha$} & \textbf{Accuracy} & \textbf{F1} \\

\midrule

\multirow{3}{*}{\makecell{LLaVA-Next\\(7B)}} 
& Baseline & - & - & - & 88.45 & 87.99 \\
& SPIN & 1$\sim$32 & 0.05 & 0.0 & 89.00 & 88.52 \\
& SPIN & 1$\sim$32 & 0.05 & 0.001 & \textbf{89.07} & \textbf{88.60} \\

\midrule

\multirow{3}{*}{\makecell{Qwen2.5-VL\\(7B)}} 
& Baseline & - & - & - & 81.64 & 77.75  \\
& SPIN & 1$\sim$20 & 0.02 & 0.0 & 83.02 & 79.80  \\
& SPIN & 1$\sim$20 & 0.02 & 0.001 & \textbf{83.05} & \textbf{79.87}  \\

\bottomrule
\end{tabular}
\vspace{-2mm}
\caption{Experimental results on Multi-turn POPE using LLaVA-Next and Qwen2.5-VL.}
\vspace{-5mm}
\label{tab:latest_models_pope}
\end{table}

\subsection{Evaluations on the Latest Models}
To further assess the accessibility and robustness of our method on state-of-the-art LVLMs, we conduct evaluations under greedy decoding on LLaVA-Next \cite{liu2024llavanext} and Qwen2.5-VL \cite{Qwen2.5-VL}. The CHAIR results are reported in Table~\ref{tab:latest_models_chair}, while the POPE results are shown in Table~\ref{tab:latest_models_pope}. SPIN consistently alleviates hallucinations in these newer models and provides improved accuracy and F1 for POPE evaluation. 

It is worth noting that, due to differences in architectures in the language backbone, Qwen2.5-VL generally benefits more from suppressing a smaller fraction of heads. Specifically, Qwen2.5-VL comprises 28 layers with 28 heads per layer, whereas LLaVA-Next, leveraging Mistral-7B, consists of 32 layers with 32 heads each. This indicates that each attention head in Qwen2.5-VL undertakes greater functional responsibility, so overly aggressive suppression of heads can undermine the model's capabilities. 



\subsection{CHAIR Visualization}
\begin{figure*}[htbp]
  \includegraphics[width=\textwidth]{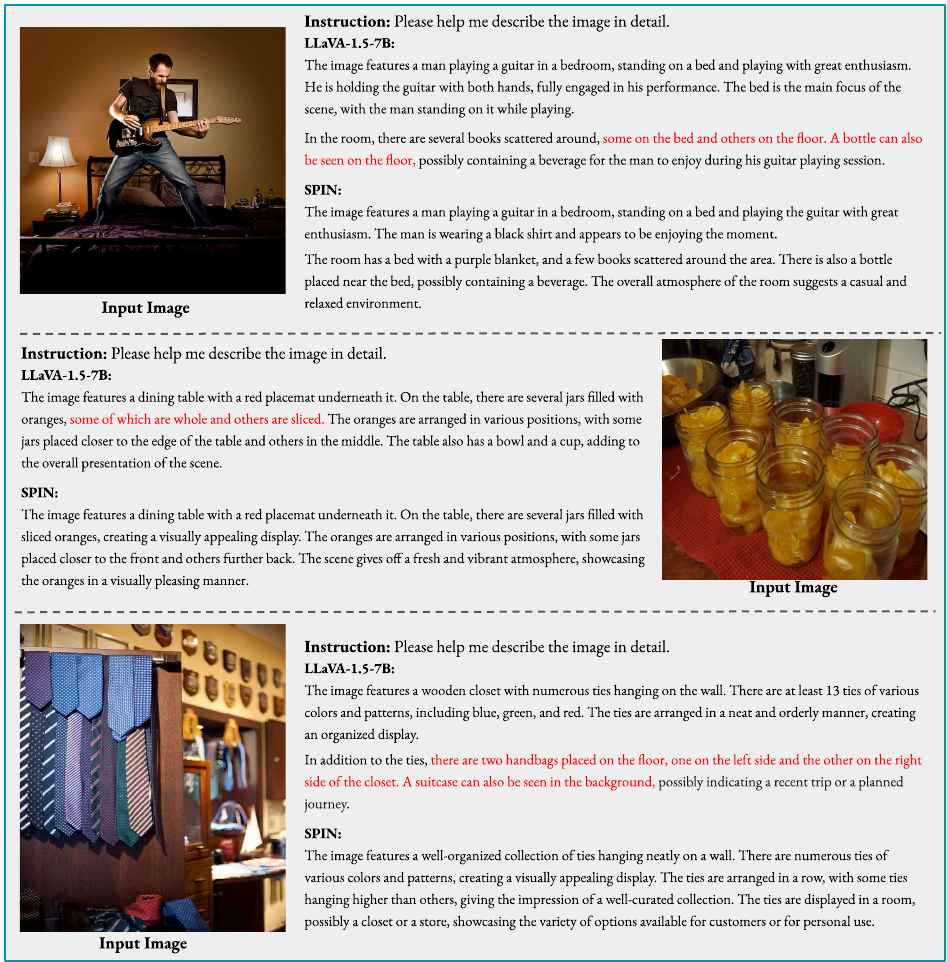}
  \caption{SPIN's effectiveness in reducing hallucinations with LLaVA-1.5 (7B).}
  \label{fig:appendix_chair_llava_7b}
\end{figure*}

\begin{figure*}[htbp]
  \includegraphics[width=\textwidth]{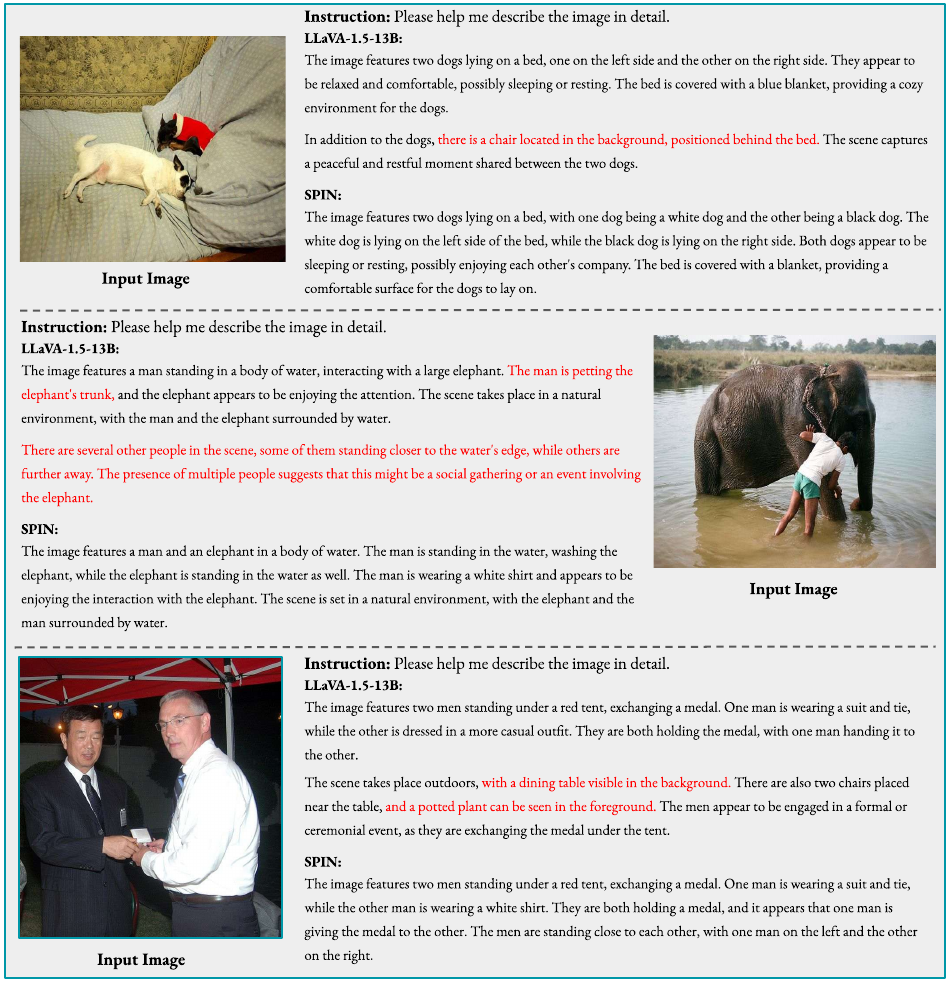}
  \caption{SPIN's effectiveness in reducing hallucinations with LLaVA-1.5 (13B).}
  \label{fig:appendix_chair_llava_13b}
\end{figure*}

\begin{figure*}[htbp]
  \includegraphics[width=\textwidth]{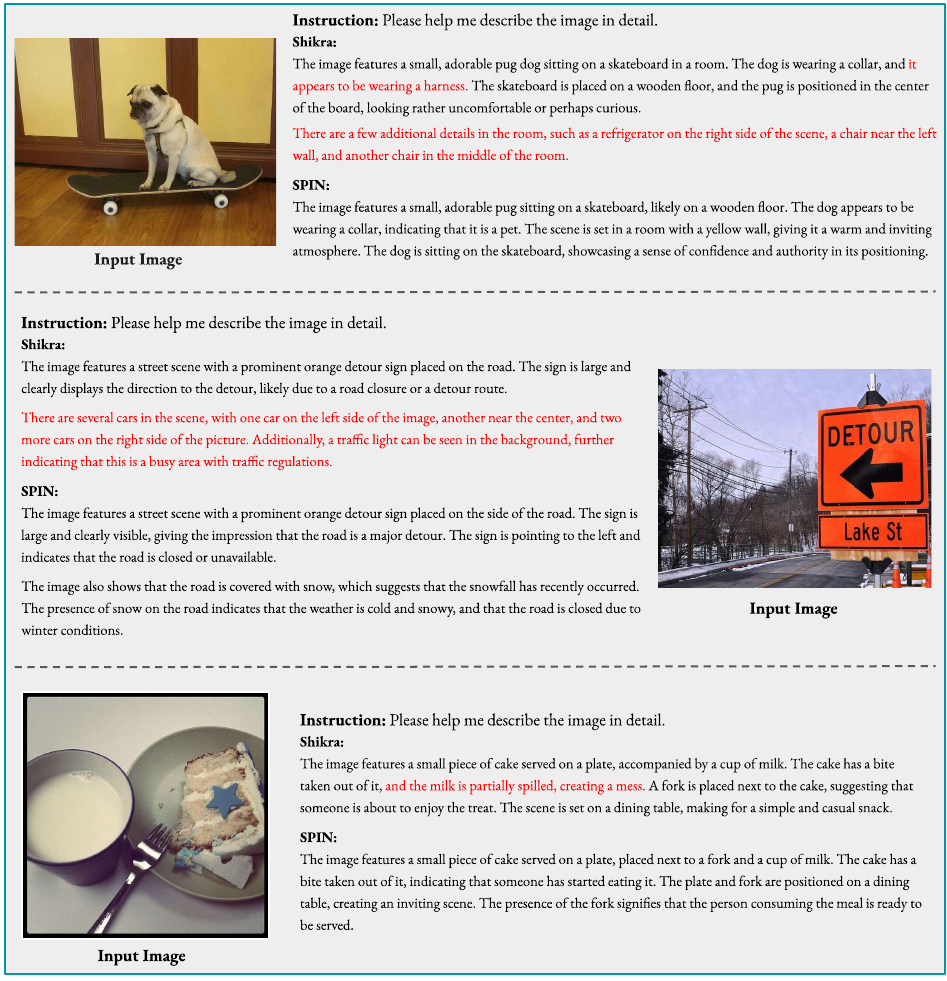}
  \caption{SPIN's effectiveness in reducing hallucinations with Shikra.}
  \label{fig:appendix_chair_shikra}
\end{figure*}

\begin{figure*}[htbp]
  \includegraphics[width=\textwidth]{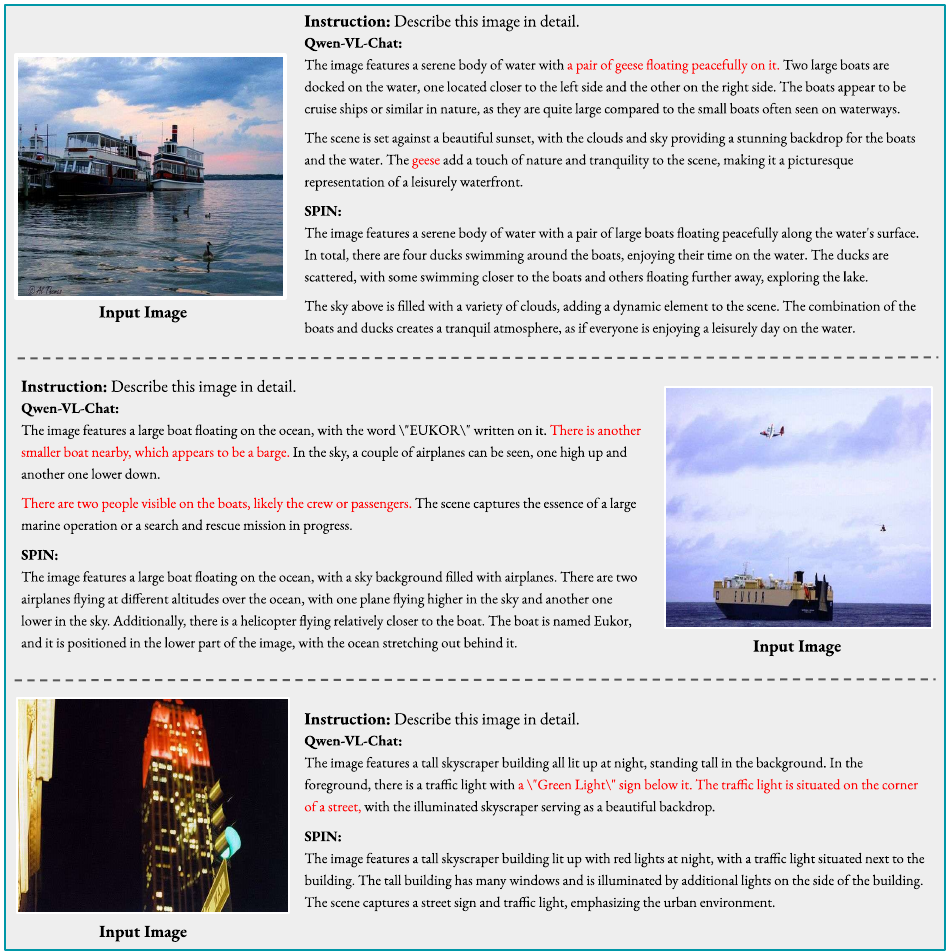}
  \caption{SPIN's effectiveness in reducing hallucinations with Qwen-VL-Chat.}
  \label{fig:appendix_chair_qwen}
\end{figure*}

\begin{figure*}[htbp]
  \includegraphics[width=\textwidth]{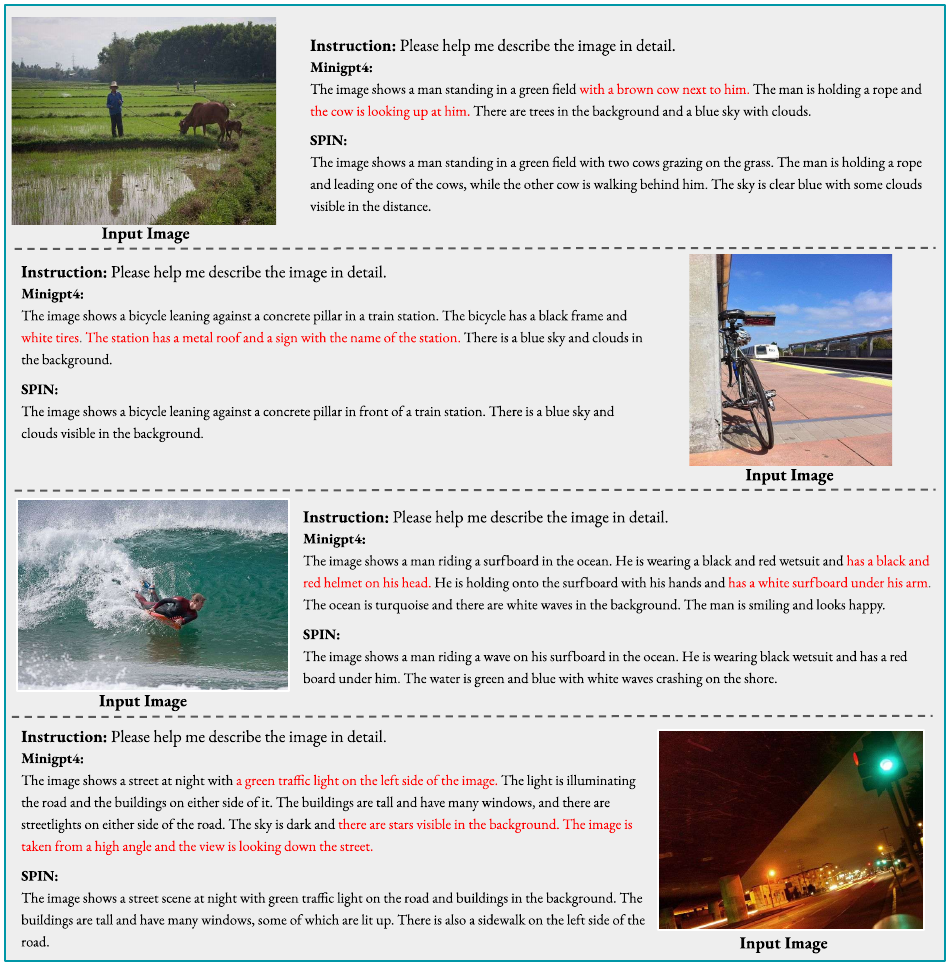}
  \caption{SPIN's effectiveness in reducing hallucinations with MiniGPT-4.}
  \label{fig:appendix_chair_minigpt}
\end{figure*}

The captions generated using the baseline and SPIN are visualized in Figures \ref{fig:appendix_chair_llava_7b}, \ref{fig:appendix_chair_llava_13b}, \ref{fig:appendix_chair_shikra}, \ref{fig:appendix_chair_qwen}, and \ref{fig:appendix_chair_minigpt}, for LLaVA-1.5 (7B), LLaVA-1.5 (13B), Shikra, Qwen-VL, and MiniGPT-4. SPIN effectively mitigates hallucinations generated by the baseline models.




\begin{table*}[tbhp]
\centering
\caption{The prompt used for GPT-4o evaluation adopted from~\citet{liu2024PAI}.}
\label{tab:gpt4o_prompt}
\begin{tabular}{p{0.95\textwidth}}
\toprule
\textbf{GPT-4o Prompt} \\
\midrule
You are required to score the performance of two AI assistants in describing a given image. You should pay extra attention to the hallucination, which refers to the part of descriptions that are inconsistent with the image content, such as claiming the existence of something not present in the image or describing incorrectly in terms of the counts, positions, or colors of objects in the image. Please rate the responses of the assistants on a scale of 1 to 10, where a higher score indicates better performance, according to the following criteria:

1: Accuracy: whether the response is accurate with respect to the image content. Responses with fewer hallucinations should be given higher scores.

2: Detailedness: whether the response is rich in necessary details. Note that hallucinated descriptions should not count as necessary details.

Please output the scores for each criterion, containing only two values indicating the scores for Assistant 1 and 2, respectively. The two scores are separated by a space. Following the scores, please provide an explanation of your evaluation, avoiding any potential bias and ensuring that the order in which the responses were presented does not affect your judgment.\vspace{0.5em}

[Assistant 1] \par
\{Response of Assistant 1\} \par
[End of Assistant 1] \par\vspace{0.5em}

[Assistant 2] \par
\{Response of Assistant 2\} \par
[End of Assistant 2]\vspace{0.5em}

Output format: 

Accuracy: \textless Scores of the two answers\textgreater \\
Reason: \vspace{0.5em}

Detailedness: \textless Scores of the two answers\textgreater \\
Reason: \vspace{0.5em} \\
\midrule
\end{tabular}
\end{table*}

